\documentclass[11pt]{article}

\usepackage[final]{acl}

\usepackage{times}
\usepackage{latexsym}

\usepackage[T1]{fontenc}
\usepackage[utf8]{inputenc}

\usepackage{microtype}

\usepackage{inconsolata}

\usepackage{graphicx}
\usepackage{times}  % DO NOT CHANGE THIS
\usepackage{helvet}  % DO NOT CHANGE THIS
\usepackage{courier}  % DO NOT CHANGE THIS
\PassOptionsToPackage{hyphens}{url} 
\usepackage{url}  % DO NOT CHANGE THIS
\usepackage{natbib}  % DO NOT CHANGE THIS AND DO NOT ADD ANY OPTIONS TO IT
\usepackage{caption} % DO NOT CHANGE THIS AND DO NOT ADD ANY OPTIONS TO IT
\usepackage{multirow}
\usepackage{booktabs}
\usepackage{algorithm}
\usepackage{algorithmic}
\usepackage{amsmath}
\usepackage{amssymb} % for checkmark and xmark
\usepackage{xcolor}  % for colors
\definecolor{olivegreen}{RGB}{85,107,47}
\newcommand{\xmark}{\ding{55}} 
\usepackage{pifont}
\usepackage{newfloat}
\usepackage{listings}
\DeclareCaptionStyle{ruled}{labelfont=normalfont,labelsep=colon,strut=off} % DO NOT CHANGE THIS
\floatstyle{ruled}
\newfloat{listing}{tb}{lst}{}
\floatname{listing}{Listing}
\usepackage{tabularx}

\title{RAGCap-Bench: Benchmarking Capabilities of LLMs in Agentic Retrieval Augmented Generation Systems}

\author{Jingru Lin$^{1\dag}$ \quad Chen Zhang$^{\dag}$ \quad \textbf{Stephen Y. Liu}\quad \textbf{Haizhou Li}$^{1,2}$     \\
  $^1$National University of Singapore, Singapore \\ 
  $^2$The Chinese University of Hong Kong, Shenzhen, China \\
        \tt \{jingrulin\}@u.nus.edu
  }
\date{}

\begin{document}
\maketitle
\def\thefootnote{\dag}\footnotetext{Equal Contribution}\def\thefootnote{\arabic{footnote}}
\begin{abstract}
Retrieval-Augmented Generation (RAG) mitigates key limitations of Large Language Models (LLMs), such as factual errors, outdated knowledge, and hallucinations, by dynamically retrieving external information. Recent work extends this paradigm through agentic RAG systems, where LLMs act as agents to iteratively plan, retrieve, and reason over complex queries. However, these systems still struggle with challenging multi-hop questions, and their intermediate reasoning capabilities remain underexplored. To address this, we propose RAGCap-Bench, a capability-oriented benchmark for fine-grained evaluation of intermediate tasks in agentic RAG workflows. We analyze outputs from state-of-the-art systems to identify common tasks and the core capabilities required for their execution, then construct a taxonomy of typical LLM errors to design targeted evaluation questions. Experiments show that “slow-thinking” models with stronger RAGCap performance achieve better end-to-end results, underscoring the benchmark’s validity and the importance of enhancing these intermediate capabilities. 

\end{abstract}

\section{Introduction}
Although LLMs demonstrate impressive performance across diverse domains~\citep{grattafiori2024llama,hadi2023survey,xi2024towards}, their reliance on internal knowledge often leads to factual errors, outdated knowledge, and hallucinations, reducing their reliability on complex and dynamic queries~\citep{mckenna-etal-2023-sources,ji2023survey,huang2025survey}. These problems are alleviated with RAG which equips the LLMs with an external knowledge base~\citep{NEURIPS2020_6b493230,izacard2023atlas,gao2023retrieval,jiang-etal-2023-active}. However, traditional RAG system  suffers from problems such as static information sources, shallow reasoning, and context integration, etc~\citep{singh2025agentic}. 
Recent progress in LLM-based agentic RAG systems presents a more promising approach to mitigate these issues and improve task handling capabilities~\citep{li2025webthinker,wu2025webdancer,zheng2025deepresearcher}. 
Agentic RAG augments the retrieval pipeline by giving the LLM greater autonomy. Instead of relying solely on static retrieval, the model actively interacts with open web environments to dynamically gather and filter information, apply structured reasoning over the retrieved context, and plan adaptively to address complex queries.

Existing evaluations of agentic RAG systems primarily rely on end-to-end multi-hop QA benchmarks~\citep{krishna-etal-2025-fact,rein2023gpqagraduatelevelgoogleproofqa,zhou2025browsecomp,xi2025infodeepseek}. While these evaluations are reliable and efficient, they offer limited fine-grained feedback on how to improve system components. Addressing this gap requires examining the intermediate processes that shape overall performance. Agentic RAG systems iteratively perform planning, retrieval, and reasoning, where errors in any stage can compound—much like mistakes in the steps of a complex math problem. To isolate these critical intermediate tasks and the capabilities they demand, we introduce RAGCap-Bench, a fine-grained benchmark for component-level evaluation within agentic RAG systems.

RAGCap-Bench comprises four core tasks distilled from an analysis of recent open-source agentic RAG systems~\citep{li2025webthinker,jin2025decoupled,li2025websailor}. These tasks capture essential model capabilities in agentic workflows—\textit{\textbf{Planning}}, \textit{\textbf{Evidence Extraction}}, \textit{\textbf{Grounded Reasoning}}, and \textit{\textbf{Noise Robustness}}. Evaluation questions for each task are derived from a taxonomy of common intermediate-stage LLM errors and are presented as multiple-choice (MCQ) items. To ensure high quality, the questions are constructed through rigorous filtering of diverse agentic trajectories, followed by thorough manual inspection and annotation.

By isolating specific skills, RAGCap-Bench enables decoupled and interpretable capability assessment while maintaining practical relevance: across diverse LLMs, benchmark performance strongly correlates with end-to-end results on complex multi-hop QA tasks (\S\ref{sec:downstream}). The benchmark is also substantially more efficient than full agentic evaluations, for example, evaluating DeepSeek-R1 on RAGCap-Bench is roughly 50× faster than running the same model through the complete WebThinker workflow~\citep{li2025webthinker} on BrowseComp-ZH~\citep{zhou2025browsecomp}, a dataset comparable in size to RAGCap-Bench. Key differences between RAGCap-Bench and existing QA benchmarks are summarized in Table~\ref{tab:comparison}.

\renewcommand{\arraystretch}{1.2}
\begin{table}[t]
\centering
% \begin{tabular}{p{2cm}|p{7cm}|p{5cm}}
\resizebox{\linewidth}{!}{
\begin{tabular}{cccc}
\hline
\textbf{Benchmark} & \textbf{Lang} & \textbf{Open Web.} & \textbf{Intermediate Skills} \\ 
 % & & \textbf{Web.} & \textbf{Eval.} \\ 
\hline
% \multicolumn{4}{c}{\textit{Reward Models Bennchmarks}} \\
% \hline
% RAG-RewardBench & En & \textcolor{red}{\xmark} & \textcolor{red}{\xmark} \\
2WikiMultiHopQA~\shortcite{xanh2020_2wikimultihop} & En & \textcolor{red}{\xmark} & \textcolor{red}{\xmark} \\
MuSiQue~\citep{trivedi2021musique} & En & \textcolor{red}{\xmark} & \textcolor{red}{\xmark} \\
% \hline
% \multicolumn{4}{c}{\textit{Agentic RAG systems Bennchmarks}} \\
% \hline 
% Frames & En & \textcolor{green}{\checkmark} & \textcolor{red}{\xmark} \\
BrowseComp\citep{zhou2025browsecomp} & En & \textcolor{green}{\checkmark} & \textcolor{red}{\xmark} \\
BrowseComp-ZH\citep{zhou2025browsecomp} & Zh & \textcolor{green}{\checkmark} & \textcolor{red}{\xmark} \\
InfoDeepSeek\citep{xi2025infodeepseek} & En\&Zh & \textcolor{green}{\checkmark} & \textcolor{red}{\xmark} \\

\hline
RAGCap-Bench (ours) & En\&Zh & \textcolor{green}{\checkmark} & \textcolor{green}{\checkmark} \\
\hline
\end{tabular}}
\caption{Comparisons between different benchmarks.}
\label{tab:comparison}
\end{table}

\section{Problem Formulation}
\label{sec:problem_formulation}

\begin{figure*}[t!]
\centering
\includegraphics[width=1.0\textwidth]{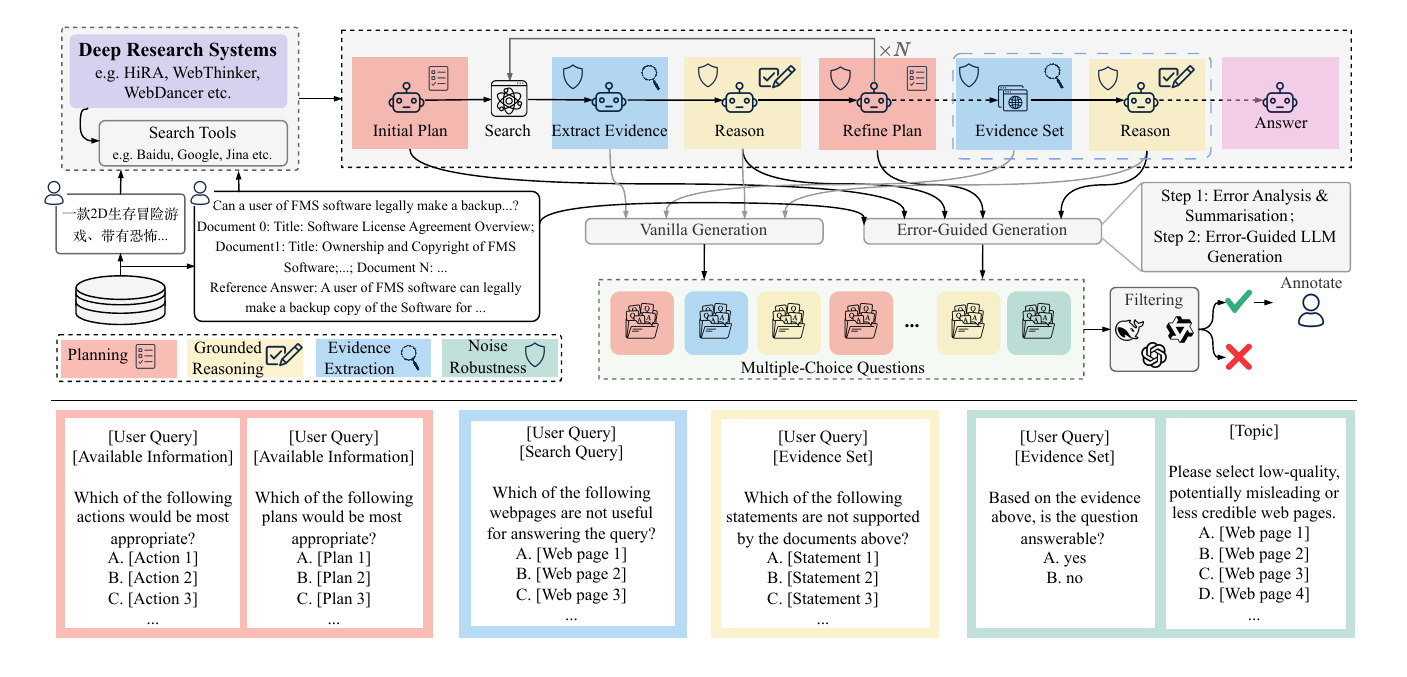}
\caption{General agentic RAG pipeline and data construction processes for RAGCap-Bench. We take queries from open-source question-answering (QA) datasets. We then run different deep research systems and collect the intermediate outputs. Two main strategies, Vanilla Generation and Error-Guided Generation are deployed to generate the MCQs based on queries and the intermediate outputs. The generated MCQs are filtered to ensure the quality and experts are recruited to provide the answers. }
\label{fig:data_construction}
\end{figure*}

% agentic system
We formalize the general pipeline of an agentic RAG system and identify the intermediate tasks that capture its core model capabilities. Figure~\ref{fig:data_construction} illustrates this overall workflow. Given a user query $q \in \mathcal{Q}$, the LLM first designs its initial question-answering plan $\pi_0$ = Plan$(q)$ about how to tackle the question. In the planning task, LLMs are required to break down the question into sub-queries, extract the question's intents, or brainstorm ideas etc., depending on the nature of the query. 
% A specific agent can do any one of these tasks or all. 
With $\pi_0$, the LLM will start to solve the problem, which requires iterative live searching, reasoning, and dynamic planning up to $T$ steps. 
% With search queries determined, the agent will start live searching and problem-solving iteratively to seek relevant information within an open environment up to $T$ steps. 
At each step $t$, the agent retrieves a set of top-K evidence $E_t = \{e_1, e_2, ..., e_k\}$ from the open web using a search engine. It then analyses the evidence set and identifies useful evidence $E'_t$ from the set. 
Based on the evidence, the LLM starts to reason and produces a reasoning chain $\mathcal{R}_t = \{r_1, r_2, ..., r_m\}$ where $m$ is the number of reasoning steps. 
With current evidence and reasoning, the LLM will then refine its plan $\pi_{t+1}$ = Plan$(q | E'_{<t}, R_{<t})$, in which it determines whether current information is enough to answer the question and makes a strategic plan for the next step. If the agent thinks the current evidence and reasoning are enough, it will stop searching and proceed to generate the answer $\hat{y}$. Depending on the specific system design, some systems incrementally gather relevant evidence through iterative search and analysis, and then perform a final reasoning step over the aggregated evidence set (as indicated by the blue dotted box in Figure~\ref{fig:data_construction}) before generating the final answer $\hat{y}$.  

% evaluate intermediate tasks -> how: LLM to critic intermediate steps; 
% but: how to do it? a) from capabilities perspective; b) what LLMs are good for this? 
% benchmark: to assess prm for agentic rag systems
The complexity of the agentic RAG system makes it hard to locate the actual problem. For example, extracting a wrong evidence could lead to incorrect reasoning, which propagates to later execution stages. In RAGCap-Bench, our goal is to introduce capability-oriented evaluations that systematically assess agentic systems at the process level. For example, how well the foundation LLMs strategizes its question-answering plan $\pi_t$, extracts useful relevant evidence $E_t'$ from $E_t$, and reasons $R_t$ faithfully with $E_t'$. Furthermore, due to the dynamic nature of the open web environments, evaluating how well LLMs handle potentially low-quality, misleading, and less credible sources of information is also important. In the next section, we will introduce the construction of each capability task in detail.

\begin{figure}[!t]
\centering
\includegraphics[width=\linewidth]{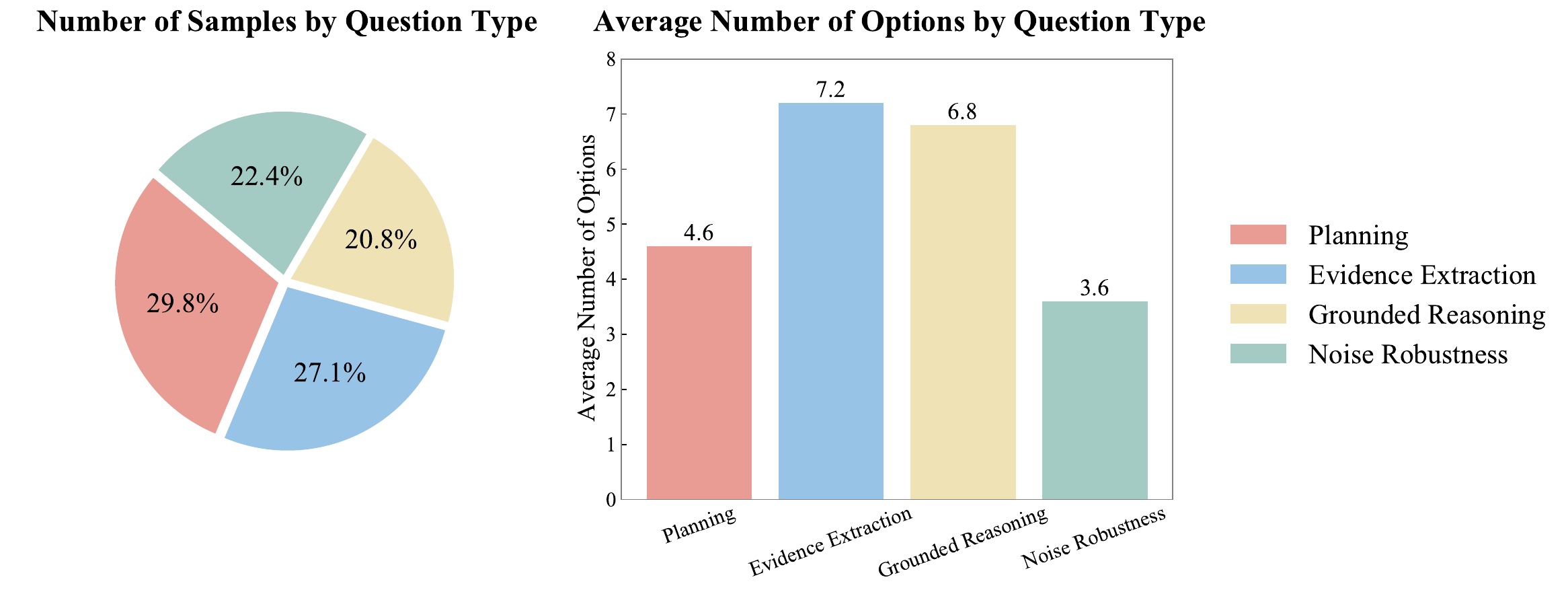}
\caption{Left: Proportion of samples in each MCQ type. Right: Average number of options in each MCQ type.}
\label{fig:dataset_stats}
\end{figure}

\section{RAGCap-Bench}
% In this section, we detail the construction of the RAGCap-Bench dataset, developed to assess the LLM's ability to identify and locate errors in the intermediate outputs of the agentic RAG systems. The questions in RAGCap-Bench are all in multiple-choice question (MCQs) format. 
Our benchmark comprises 255 curated questions. Figure~\ref{fig:dataset_stats} shows the sample distribution by task types. To construct the dataset, we draw queries from multiple open-source deep research benchmarks~\citep{zhou2025browsecomp,xi2025infodeepseek,krishna-etal-2025-fact,chen2025xbenchtrackingagentsproductivity} spanning diverse domains, including entertainment, sports, arts, technology, and medicine. Since the open-source datasets consist solely of end-to-end query-answer pairs, we utilize various agentic RAG systems to process the queries and collect execution trajectory logs. From these logs, we extract the relevant intermediate information, which is then used to generate task-specific MCQs. More details about the open-source datasets and agentic systems can be found in Appendix~\ref{appendix:datasets}.

\subsection{MCQ Format}
% Figure~\ref{} shows the general format for each type of MCQs. 
Motivated by prior works~\cite{ross2025when2call,Hendrycks2020MeasuringMM,zellers2019hellaswag}, we frame RAGCap-Bench questions as MCQs.  Each MCQ consists of a query, some intermediate outputs from the agentic RAG systems, and an instruction to ask the LLMs to choose the correct/incorrect options from all options given. The general formats of the MCQs are shown in the bottom of Figure~\ref{fig:data_construction}.
% \textcolor{red}{Each MCQ consists of a query, contexts about the intermediate task that the agentic RAG systems will perform, and an instruction prompting the LLM to identify the correct/incorrect options.}
% This prompts the LLMs to critically assess each option. 
Examples of MCQs are shown in Appendix E.

\renewcommand{\arraystretch}{1.2} % Slightly smaller row height
\begin{table*}[t]
\centering
\small % Reduce font size
\setlength{\tabcolsep}{5pt} % Reduce horizontal padding
\begin{tabularx}{\textwidth}{p{2.2cm}|X|X}
\toprule
\textbf{Question Type} & \textbf{Targeted Capability} & \textbf{Common Mistakes} \\ 
\midrule
\multirow{5}{=}{Planning} 
  & \multirow{5}{=}{The capability to interpret and break down the problem, implement an optimal/strategic problem-answering plan, and dynamically refine its plan based on current information.} 
  & Poor question interpretation \\ \cline{3-3}
  & & Omission of necessary constraints \\ \cline{3-3}
  & & Poor planning logic \\ \cline{3-3}
  & & Poor dynamic planning \\ \cline{3-3}
  & & Limited scope and depth \\
\midrule
\multirow{2}{=}{Evidence Extraction} 
  & \multirow{2}{=}{The capability to identify useful evidence from a large amount of retrieved documents.} 
  & Shallow keyword matching \\ \cline{3-3}
  & & Fail to recognise implicit connections \\ 
\midrule
\multirow{3}{=}{Grounded Reasoning} 
  & \multirow{3}{=}{The capability to reason with grounding and generate a well-supported statement.} 
  & Hallucinated support \\ \cline{3-3}
  & & Contradiction with retrieved information \\ \cline{3-3}
  & & Failed to identify implicit reasoning gap \\ 
\midrule
\multirow{2}{=}{Noise Robustness} 
  & \multirow{2}{=}{The capability to detect low-quality, less-reliable information and the capability to abstain.} 
  & Forced/Incorrect reasoning based on noisy information \\ \cline{3-3}
  & & Unable to detect low-quality, misleading, and less credible sources \\
\bottomrule
\end{tabularx}
\caption{Question types, the targeted capabilities, and common mistakes made by LLMs.}
\label{tab:error_definitions}
\end{table*}

\subsection{Dataset Construction}
There are four MCQ task types: \textit{\textbf{Planning}}, \textit{\textbf{Evidence Extraction}}, \textit{\textbf{Grounded Reasoning}} and \textit{\textbf{Noise Robustness}}. 
% Each reflects a capability required in the agentic RAG setting. 
% Figure~\ref{fig:data_construction} shows the typical pipeline for an agentic RAG system and the strategies for the construction of our dataset. 
The two main MCQs generation strategies include 1) Vanilla Generation; 2) Error-Guided Generation. The former involves directly extracting and reformatting relevant information from execution logs, framing it as MCQs to evaluate the corresponding model capabilities. The latter involves dedicated error-guided prompts to instruct the LLMs to generate high-quality and challenging MCQs. We mainly use GPT-4o~\cite{hurst2024gpt}, Qwen-Plus~\cite{yang2025qwen3} and DeepSeek-V3~\cite{liu2024deepseek} for our dataset construction.
Below shows how the MCQs of each type are constructed: 

\textit{\textbf{Planning}}: 
% As shown in Figure~\ref{fig:data_construction}, two main stages in the agentic RAG systems require planning capability. The first stage is when LLMs receive the question and start interpreting it. Based on its understanding of the user's intention, it will design a strategic question-answering plan to tackle the query. 
% The second stage comes after the LLMs have obtained some search results and performed some reasoning. When new information is gathered, the LLMs are required to determine what has and has not been answered in previous steps, and then dynamically adjust their plans according to the current information.
As shown in Figure~\ref{fig:data_construction}, planning capability is crucial at two stages of an agentic RAG workflow. First, upon receiving the initial query, the LLM interprets the user’s intent and formulates a strategic plan to guide retrieval and reasoning. Second, after retrieving evidence and performing initial reasoning, the model reassesses its progress, identifies remaining gaps, and updates its plan based on newly acquired information. Across these stages, the model must employ both convergent and divergent planning depending on the question and the information gathered so far. Convergent planning refers to the ability to narrow the search space and advance the reasoning process efficiently. In contrast, divergent planning involves broadening the search space to increase the diversity and coverage of retrieved information, ensuring that important aspects are not overlooked. In Appendix~\ref{appendix:planning}, we show the typical use cases of convergent and divergent planning for different types of queries.

% the LLM is either required to design a converging answering plan that will guide itself to derive a final deterministic answer, or a diversified answering plan that explores multiple perspectives, interpretations or possibilities to arrive at a broader, open-ended response. 
% \textcolor{red}{use tree to explain convergent \& divergent?? Put in appendix?}

To construct MCQs that assess both convergent and divergent planning abilities, we collect relevant thinking trajectories (CoT from slow-thinking agents) corresponding to the two planning stages described above from multiple agentic RAG systems. To ensure data quality, we retain only trajectories that lead to correct final answers. Among the systems analyzed, WebThinker~\citep{li2025webthinker} and HiRA~\citep{jin2025decoupled} excel in high-level question-planning, making their traces especially suitable for this purpose. Leveraging these trajectories, we prompt LLMs to generate MCQs that reflect the underlying planning processes. Additionally, since humans excel at both convergent and divergent thinking when solving complex problems, we also collect samples from open-source datasets where human-annotated stepwise problem-solving strategies are available,~\citet{chen2025xbenchtrackingagentsproductivity} for converging planning while~\citet{du2025deepresearch} for diverging planning. We then instruct the LLMs to generate MCQs based on these strategies.
% Comparing all the systems, we find that the WebThinker~\cite{li2025webthinker} and HiRA~\cite{jin2025decoupled} pipelines highly specialize in high-level question-answering planning. Therefore, we prompt the LLMs to generate MCQs based on their planning traces. 
% On top of this, we believe that humans demonstrate excellent convergent and divergent abilities in solving complex problems. Therefore, we collect samples from open source datasets where humans' stepwise problem-solving strategies are recorded as well, and instruct the LLMs to generate MCQs based on these strategies.

% , such as professional report writing, in which they are expected to expand the scope of the topic to explore multiple perspectives, interpretations or possibilities to arrive at a broader, open-ended response. 
% and narrow down their focus to provide more in-depth insights and findings. 
% Therefore, we collect outlines from expert-written reports and generate the MCQs based on these outlines. 

Appendix~\ref{appendix:prompts} provides the detailed prompts used by LLMs to generate MCQs designed to assess the planning abilities. The above-mentioned high-quality thinking trajectories from both agentic RAG systems and humans make up the correct options in the MCQs. For wrong options, we instruct the LLMs to generate them. Specifically, we employ Error-Guided Generation, which includes the identified common mistakes made by agentic RAG systems during the planning process. The common mistakes made are listed in Table~\ref{tab:error_definitions}.

% Our initial attempts to allow LLMs to generate wrong options freely led to low-quality, trivial cases that are easily identifiable by the critic models. Such cases are also unrealistic as advanced agentic RAG systems are less prone to make such superficial mistakes.

\textit{\textbf{Evidence Extraction}}: This task evaluates whether the model can identify evidence relevant to the user query. Vanilla Generation is adopted for MCQ construction. Specifically, we collect the search queries and retrieved web pages in the execution trajectories of different agentic RAG systems and each web page is formatted into an MCQ option, as shown in Figure~\ref{fig:data_construction}. The MCQs typically include more than four options, and the model should rule out all the irrelevant evidence to the user's query. Appendix~\ref{appendix:prompts} presents the prompts for the MCQs generation.

\textit{\textbf{Grounded Reasoning}}: This task evaluates whether a model can produce logical and coherent reasoning grounded in the available evidence. We employ two MCQ construction strategies: Error-Guided Generation and Vanilla Generation. For Error-Guided Generation, we use intermediate reasoning traces that lead to the correct final answer. The prompt includes the query, the evidence set, the corresponding reasoning steps, and the taxonomy of grounded-reasoning errors defined in Table~\ref{tab:error_definitions}. LLMs are instructed to generate MCQs by intentionally introducing errors into selected reasoning statements. To more realistically reflect errors made in real agentic RAG workflows, we also leverage reasoning traces that lead to incorrect final answers. Since these traces already contain mistakes, we apply Vanilla Generation, prompting the LLMs to extract diverse statements—without injecting additional errors—from the flawed reasoning. Prompts for both strategies are provided in Appendix~\ref{appendix:prompts}. Each resulting MCQ includes an instruction asking the model to identify the incorrect statements, enabling systematic evaluation of its ability to detect faulty reasoning.

\textit{\textbf{Noise Robustness}}:
Real-time search results may include noise, such as low-quality, less credible, or entirely irrelevant information. During inspection of the execution logs of different agentic RAG systems, we find that they are prone to forcing their reasoning based on the retrieved evidence, even though they may contain lots of noise. In this task, we mainly assess whether the models are robust against noisy search results. For MCQs of this task, the model is provided with search results that contain no or very limited useful information, and is prompted with two options: answer the question or admit that the question cannot be answered with the given search results. In this type of question, we deem an LLM that admits the questions are not answerable as a noise-robust one. We term the capability as noise-abstain.

% While in agentic RAG systems, the systems have to provide some answers when search limits are hit, it is important for them not to be swing by all the irrelevant information and point out that they do not have adequate information to answer the questions and are providing some reasoning based on available information. 

% Noise robustness is also important in the intermediate stages where LLMs should be resistant to noisy information so as to avoid generating incorrect reasoning and propagating it to the next stage.

% Therefore, here we want to assess if they can be robust to the search results. 

We also evaluate whether LLMs can identify less credible sources. Although useful information can sometimes be extracted from such sources, their reliability is difficult to guarantee—an especially critical concern in high-stakes domains such as medicine, law, academia, and public policy. The ability to recognize and prioritize credible sources is also valuable when multiple references present conflicting information. To assess this capability, which we term noise reliability, we use Vanilla MCQ Generation to construct questions that require models to identify the less credible sources among the provided options.
% Unlike RAG-Reward Bench, where LLM is used to synthesize counterfactual documents for assessing robustness, here we use real-world search results. 

\subsection{Dataset Post Processing}
Despite using Error-Guided Generation, LLMs may still produce trivial MCQs or exhibit issues such as hallucination and poor instruction following. To ensure question quality, we apply both difficulty filtering and format filtering. Difficulty filtering removes trivial items by prompting multiple models to answer each MCQ and discarding those for which all models choose the same option. Format filtering eliminates poorly formatted or ambiguous questions through rule-based checks and manual verification. In total, we collected 5-6k agentic trajectories from multiple systems and datasets, from which MCQs were generated. The difficulty filtering removes approximately 70-85\% of the generated items for each dataset, as many of the generated MCQs are too easy. Format filtering removes an additional 10-15\% for each dataset. This leaves a small, high-quality subset for the annotators.
All retained MCQs are annotated by human experts with access to advanced research tools. Three NLP specialists conducted the annotations, achieving an inter-annotator agreement of 0.7825 (average pairwise Jaccard similarity). Ground-truth answers were determined by majority vote; when no majority emerged, disagreements were resolved through discussion.

\subsection{Evaluation Metrics}
We adopt both Exact Match (EM) and instance-wise macro F1 score as the evaluation metrics. \textit{\textbf{Planning}} questions contain two sub-categories: converging and diverging. \textit{\textbf{Noise Robustness}} also contains two sub-categories: abstain and reliability. The EM and F1 scores are first averaged within the group. The overall score is the average of four scores from each question type. 

\section{Evaluations}
\subsection{Setups}
We evaluate a range of fast-thinking LLMs, including Qwen2.5-72B-Instruct~\citep{qwen2.5}, Qwen-Plus w/o think~\citep{yang2025qwen3}, DeepSeek-V3~\citep{liu2024deepseek}, GPT-4o~\citep{hurst2024gpt} and Gemini-1.5-Pro~\citep{team2024gemini}, and slow-thinking LLMs, including Qwen3-8B~\cite{yang2025qwen3}, Qwen3-32B w/ think~\citep{yang2025qwen3}, QwQ-32B~\citep{yang2025qwen3}, Qwen3-235B-A22B w/ think~\citep{yang2025qwen3}, DeepSeek-R1~\citep{guo2025deepseek}, O1-mini~\cite{jaech2024openai} and Gemini-2.5-Flash~\citep{comanici2025gemini}.

\renewcommand{\arraystretch}{1.1}
\begin{table*}[t]
\small
\setlength{\tabcolsep}{4pt}
\resizebox{\textwidth}{!}{
\begin{tabular}{ccccccccccccc}
\toprule
\textbf{Model} & \multicolumn{3}{c}{Planning} & \multicolumn{2}{c}{Evidence Extraction} & \multicolumn{2}{c}{Grounded Reasoning} & \multicolumn{3}{c}{Noise Robustness} & \multicolumn{2}{c}{Overall} \\ 
& EM$_c$ & F1$_c$ & EM$_d$ & EM & F1 & EM & F1 & EM$_a$ & EM$_r$ & F1$_r$ & EM & F1 \\ 
\midrule
\multicolumn{13}{c}{\textit{fast-thinking models}} \\
\midrule
Qwen2.5-72B-Instruct & 27.45 & 54.88 & 76.00 & 31.88 & \underline{79.47} & 35.85 & 77.98 & 64.86 & 10.00 & 77.18 & 39.22 & 72.37 \\
Qwen-Plus w/o think & 49.02 & 74.10 & 84.00 & 27.54 & 74.44 & 35.85 & 80.82 & 64.86 & 25.00 & 74.45 & 43.71 & 75.95 \\
Deepseek-v3 & 49.02 & 74.97 & 80.00 & 15.94 & 75.07 & 41.51 & 81.19 & 78.38 & 25.00 & 81.01 & 43.41 & \underline{78.06} \\
GPT-4o & 35.29 & 59.48 & 72.00 & 34.78 & 78.27 & 50.94 & \underline{81.89} & \underline{97.30} & 10.00 & 65.96 & 48.26 & 71.40 \\
GPT-5.5 & \underline{62.75} & \underline{80.20} & \underline{96.00} & 40.58 & 73.47 & 49.06 & 71.54 & 70.27 & \underline{35.00} & \underline{83.88} & 55.41 & 77.27 \\
Gemini-1.5-Pro & 49.02 & 75.82 & 76.00 & 18.84 & 75.37 & 37.74 & 80.49 & 54.05 & 10.00 & 73.36 & 37.78 & 76.26 \\
Gemini-3.1-Pro & 43.14 & 57.24 & 96.00 & \underline{42.02} & 76.53 & \underline{52.83} & 79.18 & 94.60 & \underline{35.00} & 82.54 & \underline{57.31} & 73.80 \\
\midrule

\multicolumn{13}{c}{\textit{slow-thinking models}} \\
\midrule
Qwen3-8B & 47.12 & 69.13 & 62.67 & 37.68 & 75.09 & 37.74 & 80.50 & 55.86 & 31.67 & 81.32 & 43.59 & 76.51 \\
Qwen3-32B w/ think & 45.10 & 70.63 & 78.67 & 25.60 & 56.49 & 54.09 & 86.97 & 57.66 & \underline{40.00} & 80.61 & 47.60 & 73.68 \\
QwQ-32B & 35.29 & 70.43 & 80.00 & 36.23 & 77.73 & 53.46 & 85.08 & 63.96 & 35.00 & \underline{81.67} & 49.21 & 78.73 \\
Qwen3-235B-A22B w/ think & 49.67 & 71.17 & 84.00 & \underline{40.10} & 75.89 & \underline{57.23} & \underline{89.29} & 56.76 & 38.33 & 82.27 & 52.93 & 79.66 \\
DeepSeek-R1 & 52.94 & 75.45 & \underline{86.67} & 34.78 & \underline{87.75} & 55.35 & 87.52 & 73.86 & 35.00 & 81.65 & \underline{53.59} & \underline{81.05} \\
O1-mini & 36.60 & 67.60 & 64.00 & 30.34 & 74.43 & 39.62 & 71.99 & 60.36 & 20.00 & 73.81 & 40.13 & 71.96 \\
Gemini-2.5-Flash & \underline{58.82} & \underline{78.41} & 82.67 & 34.78 & 76.74 & 38.36 & 80.28 & \underline{77.47} & 25.00 & 75.60 & 48.78 & 77.75 \\
\bottomrule
\end{tabular}}
\caption{Performance of different LLMs using informative prompts. EM$_c$ and F1$_c$ denote the scores for converging ability. EM$_d$ denotes the EM score for diverging ability. EM$_a$ denotes the EM score for noise-abstain. EM$_r$ and F1$_r$ denote the scores for noise-reliability. The scores are presented as percentages for clarity.}
\label{tab:results_informative}
\end{table*}

\subsection{Benchmark Results}
\label{sec:benchmark_results}
We evaluate the performance of fast-thinking and slow-thinking models across the different task types using both bare and informative prompts. All prompts are shown in Appendix~\ref{appendix:bare-and-informative}. Bare prompts refer to those with no error examples included. The LLMs are given the query, MCQs, and simple instructions about the output format. Informative prompts additionally include the few-shot examples about the common mistakes listed in Table~\ref{tab:error_definitions}. The latter shows improved and more robust performance. Table~\ref{tab:results_informative} shows the results for using informative prompts. The results are averaged over three runs. The comparison with bare prompts is shown in Section~\ref{sec:bare_prompt}.
% The results for bare prompts are shown in Appendix~\ref{appendix:bare_prompt}.

\noindent\textit{\textbf{Planning}}: We report convergent and divergent planning scores separately in Table~\ref{tab:results_informative}. EM$_c$ and F1$_c$ correspond to convergent planning, while divergent planning is evaluated using EM$_d$ only. A high EM score indicates that the model consistently selects the optimal solution path over non-optimal alternatives. In agentic RAG systems, strong EM performance translates to more efficient workflows, avoiding redundant searches and unnecessary evidence collection. In contrast, a high F1 score suggests that the model sometimes selects non-optimal paths but can still reach the correct answer—reflecting the iterative nature of agentic reasoning, where redundant steps may still lead to successful outcomes. Within the fast-thinking group, GPT-5.5 achives the strongest overall performance, substantially outperforming other models. Among slow-thinking models, Gemini-2.5-Flash lead with an EM$_c$ of 58.82\% and a F1$_c$ of 78.41\%.

\noindent\textit{\textbf{Evidence Extraction}}:
% EM: select all relevant evidences -> higher EM -> able to identify all relevant evidence, provide a strong foundation for faithful and accurate reasoning 
% f1: a) miss out repetitive evidences -> still able to extract relevant information for the next step reasoning; b) miss out key information; -> might not be able to reason & require extra planning step (additional search etc)
% c) select irrelevant data -> make the models less robust (see robustness ability)
% informative prompt: add errors + assume web browsing function if link is given, ask the model to determine if browsing the web page content will give them information to answer the question
% critic on the usefulness of evidence -> a good critique model can assess if the systems have identified useful evidence 
A high EM score in evidence extraction reflects an LLM’s ability to precisely identify relevant evidence while filtering out irrelevant information—crucial for enabling faithful and accurate reasoning in agentic RAG systems. As shown in Table~\ref{tab:results_informative}, most fast- and slow-thinking models perform poorly on this task, with EM scores below 40\%, indicating that current LLMs struggle to process information from dynamic, open-web environments. The strongest results are 42.02\% for Gemini-3.1-Pro
% 34.78\% for GPT-4o
(fast-thinking) and 40.10\% for Qwen3-235B-A22B w/ think (slow-thinking). This weakness likely stems from the mismatch between pretraining data and noisy, unstructured web content, as well as challenges in aggregating evidence across heterogeneous sources.

However, in agentic RAG settings, multiple sources often provide overlapping information, meaning that omitting some redundant evidence has little impact on downstream reasoning. Our analysis reveals three common failure types in evidence extraction: (a) omission of repetitive evidence, (b) omission of key information, and (c) inclusion of irrelevant information. While (a) is largely harmless, the impact of (b) and (c) depends on the system’s planning and noise-robustness capabilities. We elaborate these interdependencies later in this section. In this context, F1, which rewards partial correctness, captures whether a system can extract at least some of the relevant information needed to continue reasoning. As shown in Table~\ref{tab:results_informative}, most models achieve F1 scores above 70\%. DeepSeek-R1 performs best overall, reaching an F1 score of 87.75\%.

% However, in the context of agentic RAG systems, it is common for multiple sources to provide overlapping information. In such cases, the omission of some repetitive evidence has a trivial impact on the reasoning. 
% % This makes EM a harsh indicator in terms of evidence extraction. 
% Through analysis on the performance of LLMs on this question type, we summarise three common failure cases: a) omission of repetitive evidence; b) omission of key information; c) inclusion of irrelevant information. While a) is a trivial mistake, the effects of b) and c) are dependent on the system's planning and noise robustness capabilities. We will elaborate on these interdependencies later in this section. In this sense, F1, which provides partial correctness, captures whether the systems are able to extract at least some relevant information to carry on with their reasoning process.  
% From Table~\ref{tab:results_informative}, we can see the LLMs are generally able to achieve F1 scores of more than 70\%. Among all models, DeepSeek-R1 achieves the best F1 score of 81.34\% among all models.

\noindent\textit{\textbf{Grounded Reasoning}}:
% EM: this might be a better indication of this capability as the agent system requires flawless reasoning across stages 
% f1: LLMs might be susceptible to wrong reasoning, and the errors can propogate
While an F1 score reflects partial correctness in reasoning statements, it does not reflect deeper reasoning failures. A mistake made by LLMs in any intermediate reasoning stage can be propagated through subsequent stages of the agentic RAG systems. 
Conversely, even perfect extraction can be undermined if the model's reasoning process is flawed.
Therefore, the EM score serves as a stronger indicator for flawless reasoning grounded in given evidence. Despite high F1 scores (generally more than 80\%) achieved by many LLMs, the EM scores are much lower. Among fast-thinking models, the best EM score of 52.83\% is obtained by Gemini-3.1-Pro.
% 50.94\% is obtained by GPT-4o.
Among slow-thinking models, the best EM score is 57.23\%, achieved by Qwen3-235B-A22B. 

\noindent\textit{\textbf{Noise Robustness}}:
% abstain: same question, different length of search 
% reliability: LLM still recognise content more than sources -> although LLMs can still extract useful information from less reliable sources, it might be susceptible to factual inaccuracies 
For noise-abstain, only the EM score, denoted as EM$_a$ is computed. Our results show that most models are noise robust. When provided with irrelevant information, most LLMs admit that the question is not answerable. For noise reliability, most LLMs struggle to assess the credibility of information sources. As shown in Table~\ref{tab:results_informative}, many models achieve high F1$_r$ scores yet have EM$_r$ values below 50\%. Some models, such as Qwen2.5-72B-Instruct, GPT-4o, and Gemini-1.5-Pro—perform particularly poorly, with EM$_r$ scores as low as 10\%. This pattern indicates a concerning tendency to trust retrieved content indiscriminately, regardless of source reliability. A closer look at model reasoning reveals that LLMs often prioritize the content of a webpage over its credibility. They frequently accept sources that appear informative even when the origin is questionable. While this behavior may be relatively harmless when different sources agree, it becomes risky in the presence of conflicting information. In such cases, human decision-making typically involves checking the authority and trustworthiness of sources before drawing conclusions, a capability current models still lack. 
\smallskip
\noindent Although we define four distinct capabilities, they do not operate independently, but interact synergistically, collectively shaping the system’s overall effectiveness. For instance, in evidence extraction and noise robustness, the negative effects of missing key information or selecting irrelevant evidence can be mitigated if the model is sufficiently noise-robust. When sources conflict, a noise-robust model can prioritize more reliable information and guide evidence selection accordingly. Similarly, planning and noise robustness are interdependent: a concise and well-structured plan reduces unnecessary searches, thereby limiting exposure to noisy or irrelevant information and easing the burden on the system’s noise-handling capability.

% For \textit{Evidence Extraction} and \textit{Grounded Reasoning}, accurate evidence extraction provides a strong foundation for faithful reasoning. 

\subsection{With vs. Without Error-Guided Prompts}
\label{sec:bare_prompt}

\begin{figure}[!t]
\centering
\includegraphics[width=0.95\linewidth]{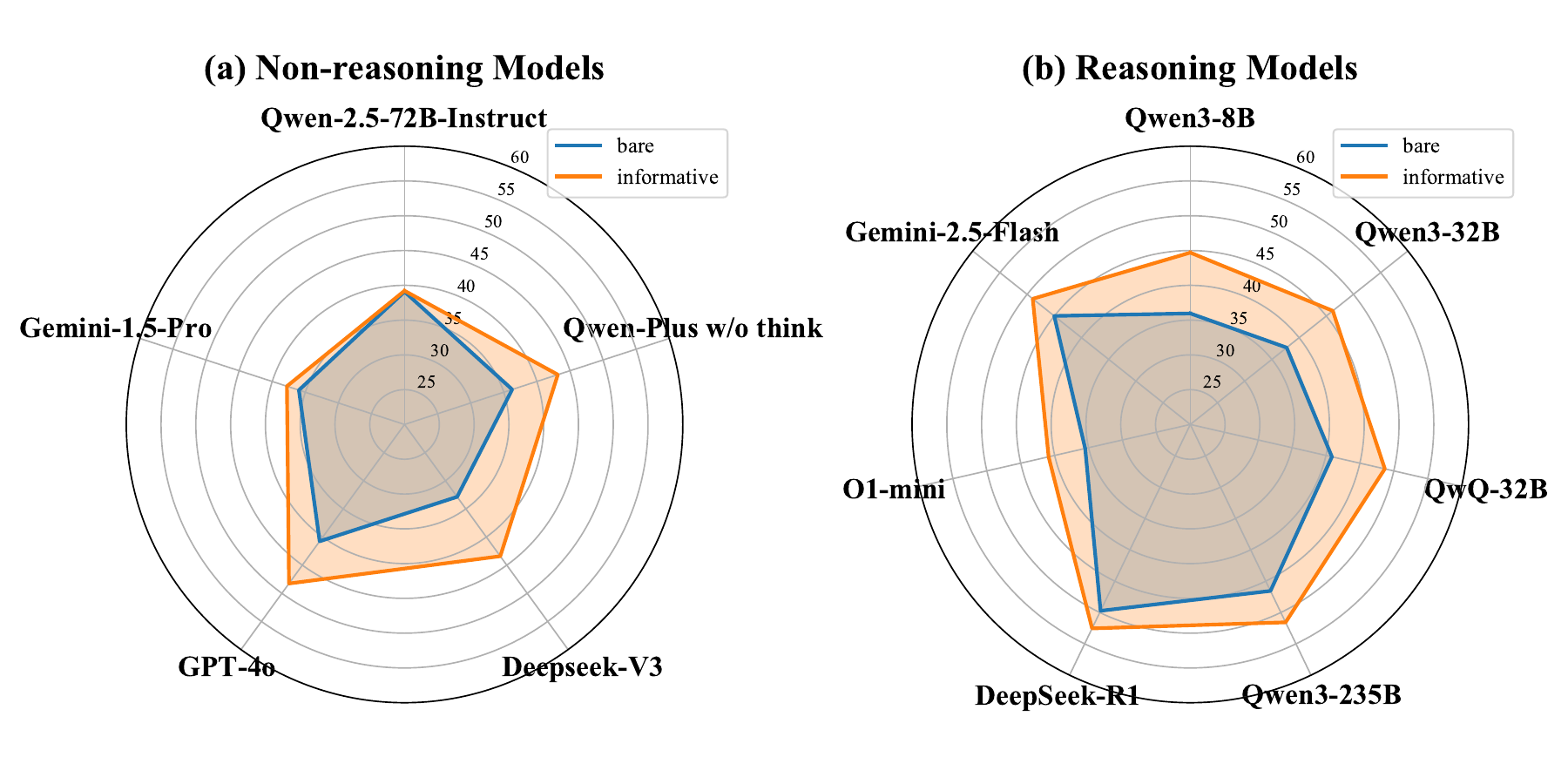}
\caption{\small RAGCap-Bench overall EM scores for different fast-thinking (left) and slow-thinking models (right), with informative (orange) and bare (blue) prompts.}
\label{fig:bare_prompt_compare}
\end{figure}

The detailed results for bare prompts (without error-guided exemplars) are provided in Appendix~\ref{appendix:bare_prompt}. Here, we compare only the overall EM scores under the two prompting settings. As shown in Figure~\ref{fig:bare_prompt_compare}, informative prompts, constructed using our error taxonomy in Table~\ref{tab:error_definitions}, consistently improve performance for both fast- and slow-thinking LLMs. This highlights a key challenge in building agentic RAG systems: even slow-thinking models struggle to engage effectively with dynamic, noisy web information without structured guidance. Consequently, developing robust agentic RAG systems often requires substantial prompt engineering or additional post-training techniques~\citep{asai2024self,zhang2025process}.

\subsection{Correlation with Downstream}
\label{sec:downstream}
\begin{figure}[!t]
\centering
\includegraphics[width=0.95\linewidth]{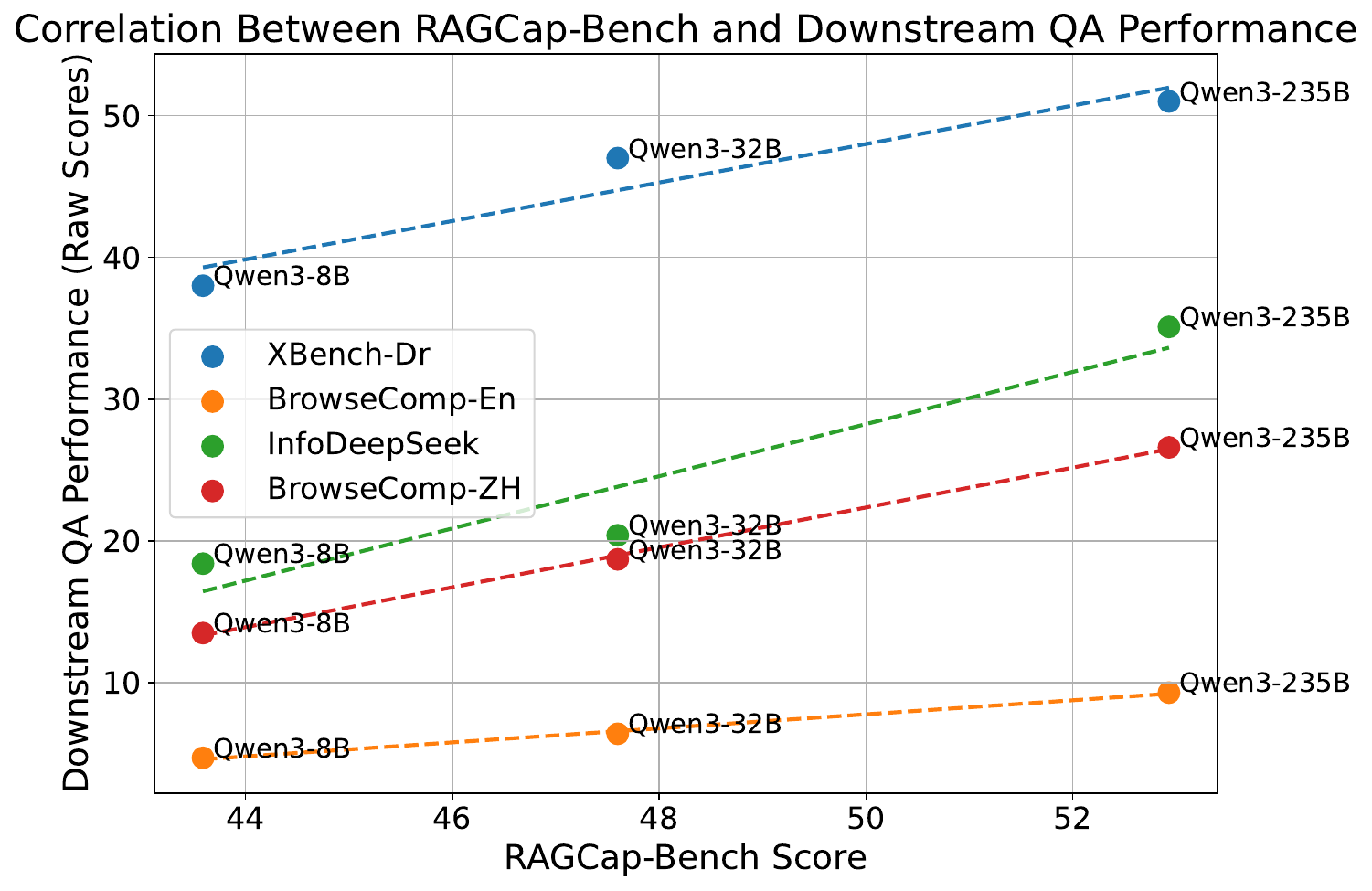}
\caption{\small Correlation of model performance on RAGCap-Bench and their respective downstream QA performance.}
\label{fig:correlation}
\end{figure}

A strong benchmark for intermediate tasks should reliably predict end-to-end performance. To evaluate this, we compare model results on RAGCap-Bench with their downstream QA accuracy. We test three LLMs of different scales (Qwen3-8B, Qwen3-32B, and Qwen3-235B-A22B) using the WebThinker workflow. Downstream performance is measured as pass@1 accuracy on BrowseComp, BrowseComp-Zh, XBench-DeepSearch, and InfoDeepSeek, with each inference limited to 10 Google Search API calls. Figure~\ref{fig:correlation} shows a clear positive correlation between RAGCap-Bench scores and the models' respective performance on each of the downstream QA benchmark when models act as agents in the WebThinker pipeline. This demonstrates that RAGCap-Bench provides a reliable and efficient proxy for more expensive and time-consuming end-to-end QA evaluations.

As shown in Figure~\ref{fig:correlation}, larger models generally perform better. This naturally raises questions of whether both RAGCap-Bench and the downstream tasks are correlated with model scale only. To inbestigate this, we select four comparably sized models from Qwen family, and evaluate them on 200 sampled questions from the four test sets shown in Figure~\ref{fig:correlation}. The results, summarised in Table~\ref{tab:similar_scale} show that neither RAGCap-Bench nor downstream task performance is solely determined by model scale. Moreoever, the strong alignment between RAGCap-Bench scores and downstream task results suggests that RAGCap-Bench remains a reliable indicator of downstream performance.

% Furthermore, the performance of Qwen3-8B and Qwen3-32B, while the former is much smaller in size, are close in RAGCap-Bench, with the overall EM scores of 44.96\% and 46.20\% respectively. This is also reflected in their downstream performance. From Figure~\ref{fig:correlation}, we can see that the performance of Qwen3-8B and Qwen3-32B is close on the four datasets for the HiRA workflow, as well as on InfoDeepSeek for the WebThinker workflow. 

% two representative agentic RAG pipelines, including WebThinker (ReACT-based) and HiRA (Multi-Agent-based), yielding a total of six configurations. 

\begin{table}[!t]
\centering
\resizebox{\linewidth}{!}{
\begin{tabular}{ccccc}
\toprule
\textbf{Models} & \multicolumn{1}{c}{\textbf{Downstream Tasks}} &  \multicolumn{2}{c}{\textbf{RAGCap-Bench}} \\ 
\cmidrule{2-4}
& Accuracy (\%) & EM & F1 \\ 
\midrule
Qwen3-30B-A3B-Base & 1.33 & 10.34 & 34.27 \\
Qwen3-30B-A3B-Thinking-2507	 & 14.00 & 35.38 & 45.51 \\
Qwen3-30B-A3B & 18.33 & 41.00 & 70.24 \\ 
QwQ-32B & 20.0 & 49.21 & 78.73 \\
\bottomrule
\end{tabular}}
\caption{The second column reports the point-biserial correlation coefficient between the evaluator scores and the downstream performance. All correlation scores are statistically significant (<0.05). The third and fourth columns show corresponding EM and F1 scores on RAGCap-Bench.}
\label{tab:similar_scale}
\end{table}

% \renewcommand{\arraystretch}{1.2}
% \begin{table}[t]
% \centering
% \resizebox{\linewidth}{!}{
% \begin{tabular}{ccccc}
% \hline
% \textbf{System} & \textbf{LLM} & \textbf{RAGCap-Bench (EM)} & \textbf{InfoDeepSeek} & \textbf{BrowseComp} \\ 
% \hline
% WebThinker & Qwen3-8B & 44.96 & 18.4 & 8.0 \\
% & Qwen3-32B & 46.20 & 20.4 & 16.3 \\
% & Qwen3-235B & 51.57 & 30.6 & 26.6 \\ 
% \hline
% HiRA & Qwen3-8B & 44.96 & 26.9 & 15.8 \\
% & Qwen3-32B & 46.20 & 28.6 & 16.3 \\
% & Qwen3-235B & 51.57 & 34.7 & 23.3 \\ 

% \hline
% \end{tabular}}
% \caption{Correlation.}
% \label{tab:downstream}
% \end{table}

\subsection{Relationship With Direct Scoring}

In this section, we examine the relationship between RAGCap-Bench and direct evaluation of intermediate reasoning steps. We randomly sample 500 inference trajectories from the WebThinker experiments described in Section~\ref{sec:downstream} and apply an LLM-as-a-judge setup, prompting various LLMs to score grounded reasoning (thought–action) and evidence extraction (observation–extraction) on a 1–10 scale at each step. Scores are then averaged across the full reasoning trajectory. The evaluation prompts are provided in Appendix~\ref{appendix:critic_prompts}. Because ground-truth human scores are unavailable, we approximate evaluation quality using the point-biserial correlation coefficient~\cite{tate1954correlation} between the LLM-assigned overall scores and binary downstream outcomes (correct vs. incorrect final answers). We assess three LLMs of different sizes: Qwen3-8B, Qwen3-32B, and Qwen3-235B. Results are shown in Table~\ref{tab:critic_model}. The observed correlations align closely with each model’s performance on the evidence-extraction and grounded-reasoning categories of RAGCap-Bench. Qwen3-235B achieves the strongest correlation and also the highest EM scores on RAGCap-Bench. Qwen3-8B and Qwen3-32B exhibit weaker correlations, consistent with their lower EM scores on these tasks. Notably, Qwen3-8B shows a  higher correlation than Qwen3-32B on evidence extraction, which is also supported by its higher EM and F1 scores in RAGCap-Bench.

\begin{table}[t]
\centering
\resizebox{\linewidth}{!}{
\begin{tabular}{ccccc}
\toprule
\textbf{Evidence Extraction} & \multicolumn{1}{c}{\textbf{Direct Scoring}} &  \multicolumn{2}{c}{\textbf{RAGCap-Bench}} \\ 
\cmidrule{2-4}
& Correlation & EM & F1 \\ 
\midrule
Qwen3-8B & 0.210 & 37.68 & 75.09 \\
Qwen3-32B & 0.113 & 25.60 & 56.49 \\
Qwen3-235B & 0.528 & 40.10 & 75.89 \\ 
\midrule
\textbf{Grounded Reasoning} & \multicolumn{1}{c}{\textbf{Direct Scoring}} &  \multicolumn{2}{c}{\textbf{RAGCap-Bench}} \\ 
\cmidrule{2-4}
& Correlation & EM & F1 \\ 
\midrule
Qwen3-8B & 0.291 & 37.74 & 80.50 \\
Qwen3-32B & 0.316 & 54.09 & 86.97 \\
Qwen3-235B & 0.338 & 57.23 & 89.29 \\ 
\bottomrule
\end{tabular}}
\caption{The second column reports the point-biserial correlation coefficient between the evaluator scores and the downstream performance. All correlation scores are statistically significant (<0.05). The third and fourth columns show corresponding EM and F1 scores on RAGCap-Bench.}
\label{tab:critic_model}
\end{table}

\subsection{Ablations}
\begin{figure}[!t]
\centering
\includegraphics[width=0.95\linewidth]{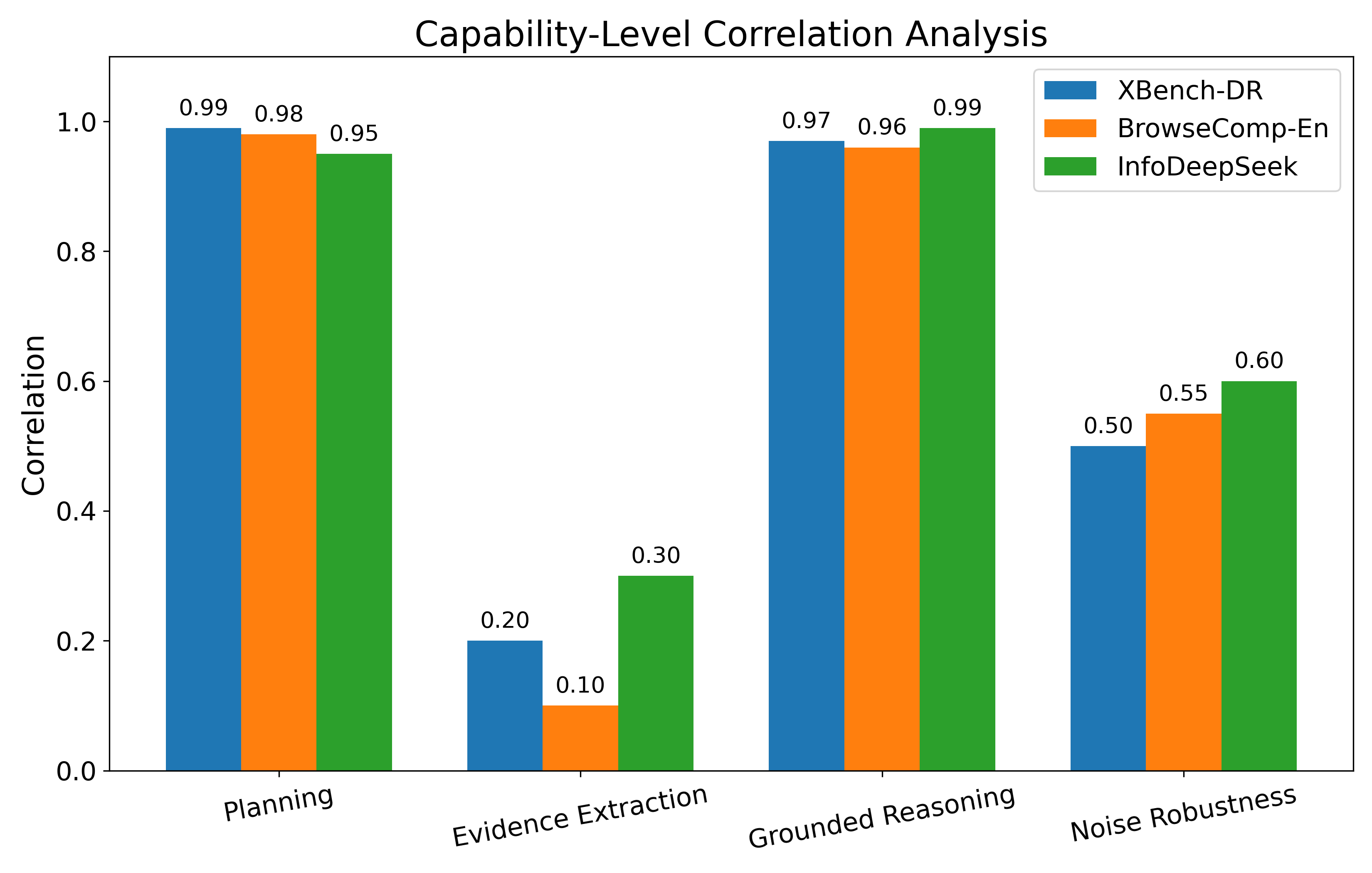}
\caption{\small Pearson correlation between sub-capability scores of model performance on RAGCap-Bench and their respective downstream QA performance.}
\label{fig:sub_cap_correlation}
\end{figure}

First, we provide a finer-grained analysis on which sub-capabilities best explain the downstream tasks. We compute the Pearson correlations between sub-capability scores and downstream performance, shown in Figure~\ref{fig:sub_cap_correlation}. Overall, we observe that grounded reasoning and planning capabilities show the strongest correlation with downstream performance, while evidence extraction exhibits weaker correlation. However, note that the capabilities do not operate independently, as we mentioned in Section~\ref{sec:benchmark_results}. They interact synergistically, collectively shaping the system's overall effectiveness.

\begin{table}[!t]
\centering

\resizebox{\linewidth}{!}{
\begin{tabular}{l l cc}
\toprule
\textbf{Model} & \textbf{Setting} & \textbf{EM} & \textbf{F1} \\
\midrule
Qwen3.5-27B & Not Shuffled & 46.14 & 81.07 \\
Qwen3.5-27B & Shuffled & 45.10 & 78.15 \\
\midrule
Qwen3-30B-A3B-Base & Not Shuffled & 10.34 & 34.27 \\
Qwen3-30B-A3B-Base & Shuffled & 10.01 & 40.60 \\
\midrule
Qwen3-30B-A3B-Thinking-2507 & Not Shuffled & 35.38 & 45.51 \\
Qwen3-30B-A3B-Thinking-2507 & Shuffled & 31.50 & 49.21 \\
\bottomrule
\end{tabular}}
\caption{Performance comparison under shuffled and non-shuffled settings.}
\label{tab:shuffle_results}
\end{table}

Next, we analyse the effects of the positions of the correct answers in the MCQ options as LLMs are known to exhibit positional bias when answering multiple-choice questions. Hence, reporting results under shuffled option orders would help ensure the conclusions are not an artifact of option placement. To verify this, we shuffle the MCQ options and compare results before and after shuffling. The results in Table~\ref{tab:shuffle_results} show that shuffling has a limited impact on overall performance.

\section{Related Work}

\textbf{Agentic RAG Systems}:
While RAG equips the LLMs with access to external knowledge bases, mitigating issues such as factual errors and hallucinations, it still struggles to solve complex tasks~\cite{singh2025agentic}. Agentic RAG has enhanced the capabilities of the traditional RAG systems. It no longer treats the LLM merely as a passive text generator, but as an active agent capable of adaptive planning, dynamic information seeking, and iterative reasoning~\citep{asai2024self,yao2023react,li2025searcho1agenticsearchenhancedlarge,feng2025airrag,zheng2025deepresearcher,pan2025multiagent}. Notably, this emerging paradigm has seen increasing adoption in practical applications, including OpenAI Deep Research~\citep{openaideepresearch}, Gemini Deep Research~\citep{geminideepresearch}, Perplexity Deep Research~\citep{perplexitydeepresearch} etc., all of which leverage LLMs as autonomous agents.

\noindent \textbf{Agentic RAG Benchmark}:
With the rise of agentic RAG systems, comprehensive and systematic benchmarking becomes increasingly important for uncovering potential weaknesses and steering the development of more effective and capable systems. Some benchmarks are proposed for this purpose~\cite{xi2025infodeepseek,zhou2025browsecomp,c}. However, most of these efforts focus on question-answering (QA) tasks that evaluate a system's ability to answer challenging, multi-hop questions. Although they are useful indicators of end-task performance, they offer limited insight into the intermediate planning, retrieving and reasoning tasks executed by the systems.

\section{Conclusion}
This work introduces RAGCap-Bench, a capability-oriented benchmark designed for fine-grained, component-wise evaluation for the agentic RAG systems. RAGCap-Bench addresses the critical gap in existing benchmarks, which lack evaluation on the intermediate processes of the systems. Experimental results demonstrate that the RAGCap-Bench scores are correlated with the downstream task performance, highlighting its practical relevance. In addition, we conduct exploratory experiments to show the potential of using LLMs to assess intermediate outputs of the agentic RAG systems.
This paves the way for future research into the integration of LLMs as a means of improving the agentic RAG systems. 

\section*{Limitation}

RAGCap-Bench is limited in size, with approximately 60 questions per category. This relatively small scale may reduce statistical power and makes it difficult to draw strong conclusions from small performance differences across models. Expanding the benchmark is challenging because constructing each MCQ requires running computationally expensive agentic RAG systems, manually inspecting multi-step reasoning trajectories, and carefully annotating each option, often involving full review of source webpages to assess evidence quality and credibility. In future work, we plan to explore automated pipelines to enable the construction of larger-scale MCQ datasets.

\bibliography{acl_latex}

\newpage
\appendix

\section{Datasets}
\label{appendix:datasets}

\begin{table}[t]
\centering
\resizebox{\linewidth}{!}{
\begin{tabular}{ccc}
\hline
\textbf{Dataset} & \textbf{Language} & \textbf{Number of Samples} \\
\hline
glaveai/RAG-v1 & En & 100 \\
InfoDeepSeek~\citep{xi2025infodeepseek} & Zh & 245 \\
BrowseComp-Zh~\citep{zhou2025browsecomp} & Zh & 289 \\
Frames~\citep{krishna-etal-2025-fact} & En & 230 \\
XBench~\citep{chen2025xbenchtrackingagentsproductivity} & Zh & 100 \\
Deep Research Bench~\citep{du2025deepresearch} & En\&Zh & 100 \\
\hline
\end{tabular}}
\caption{Datasets used for the construction of RAGCap-Bench. Note that the original Frames contain 824 samples, we subsample 230 to run the agentic RAG systems on these 230 samples only. In addition, the original Glaveai/RAG-v1 contains 51.4k rows, we select the first 100 rows with longest list of retrieved documents.}
\label{tab:datasets}
\end{table}

All open-source datasets used for constructing RAGCap-Bench are listed in Table~\ref{tab:datasets}. The agentic RAG systems used include WebThinker~\citep{li2025webthinker}, WebSailor~\citep{li2025websailor}, WebDancer~\citep{wu2025webdancer} and HiRA~\citep{jin2025decoupled}. We run these systems on InfoDeepSeek, BrowseComp-Zh and Frames to obtain the intermediate outputs. Glaveai/RAG-v1\footnote{https://huggingface.co/datasets/glaiveai/RAG-v1} dataset includes retrieved documents as part of its content. The remaining datasets provide human-annotated stepwise problem-solving strategies, which are used to construct planning questions. 

\section{Convergent and Divergent Capabilities}
\label{appendix:planning}
\begin{figure}[t]
\centering
\includegraphics[trim=0cm 0cm 0cm 1cm, width=1.0\linewidth]{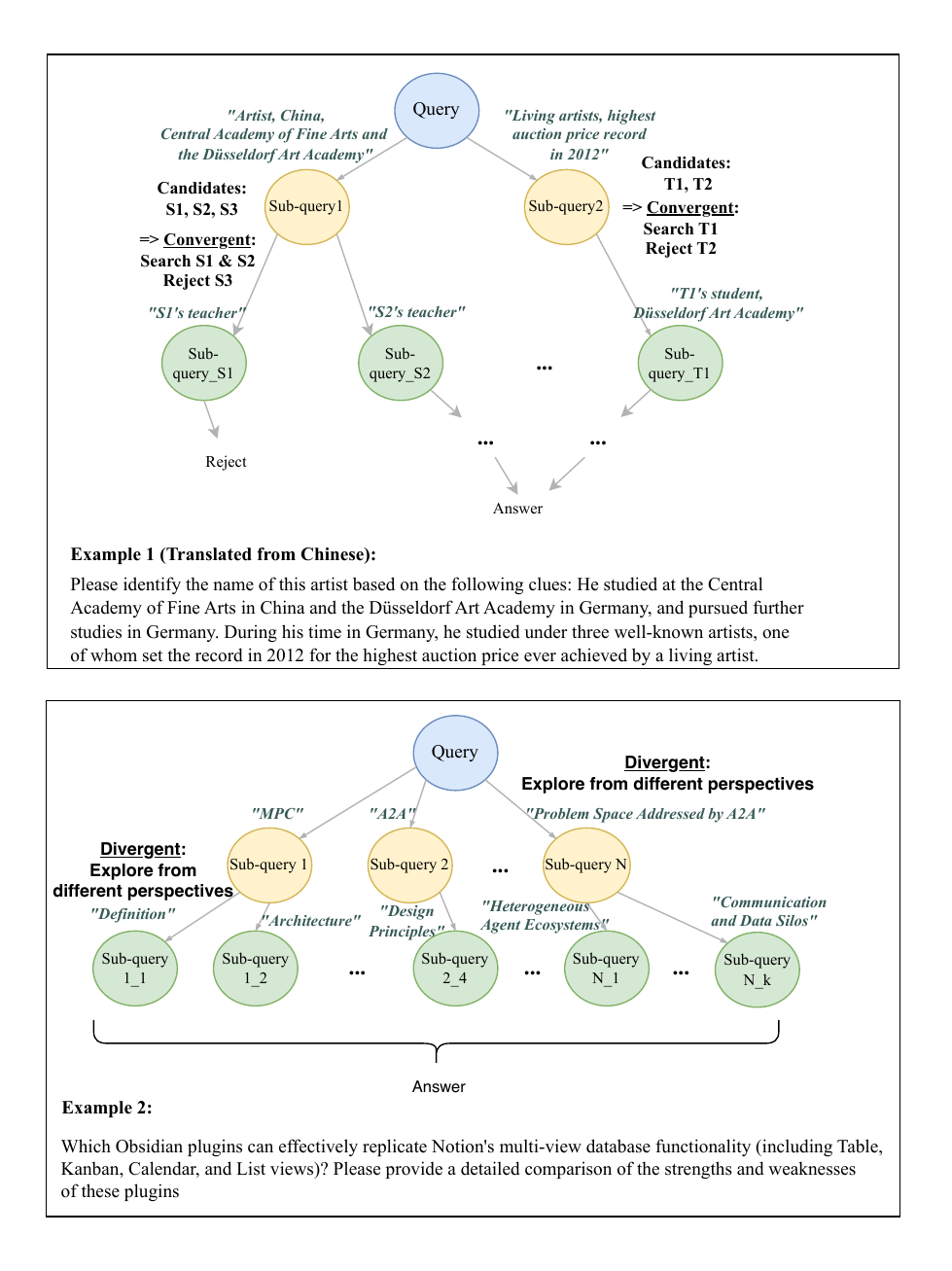}
\caption{Convergent (top) and divergent (bottom) planning capabilities required for different user queries.}
\label{fig:planning_use_cases}
\end{figure}

Figure~\ref{fig:planning_use_cases} illustrates the roles of convergent and divergent planning capabilities for different types of queries. Convergent planning capability is required to gradually narrow down the search space in order to reach a final deterministic answer. 
Divergent planning capability is required to expand a user query to explore multiple perspectives, interpretations and possibilities, so that the final answer is well-rounded and considers diverse viewpoints.

\section{Examples of Errors}
\label{appendix:error_examples}
\begin{figure*}[t]
\centering
\includegraphics[trim=0cm 3cm 0cm 0cm, width=1.0\textwidth]{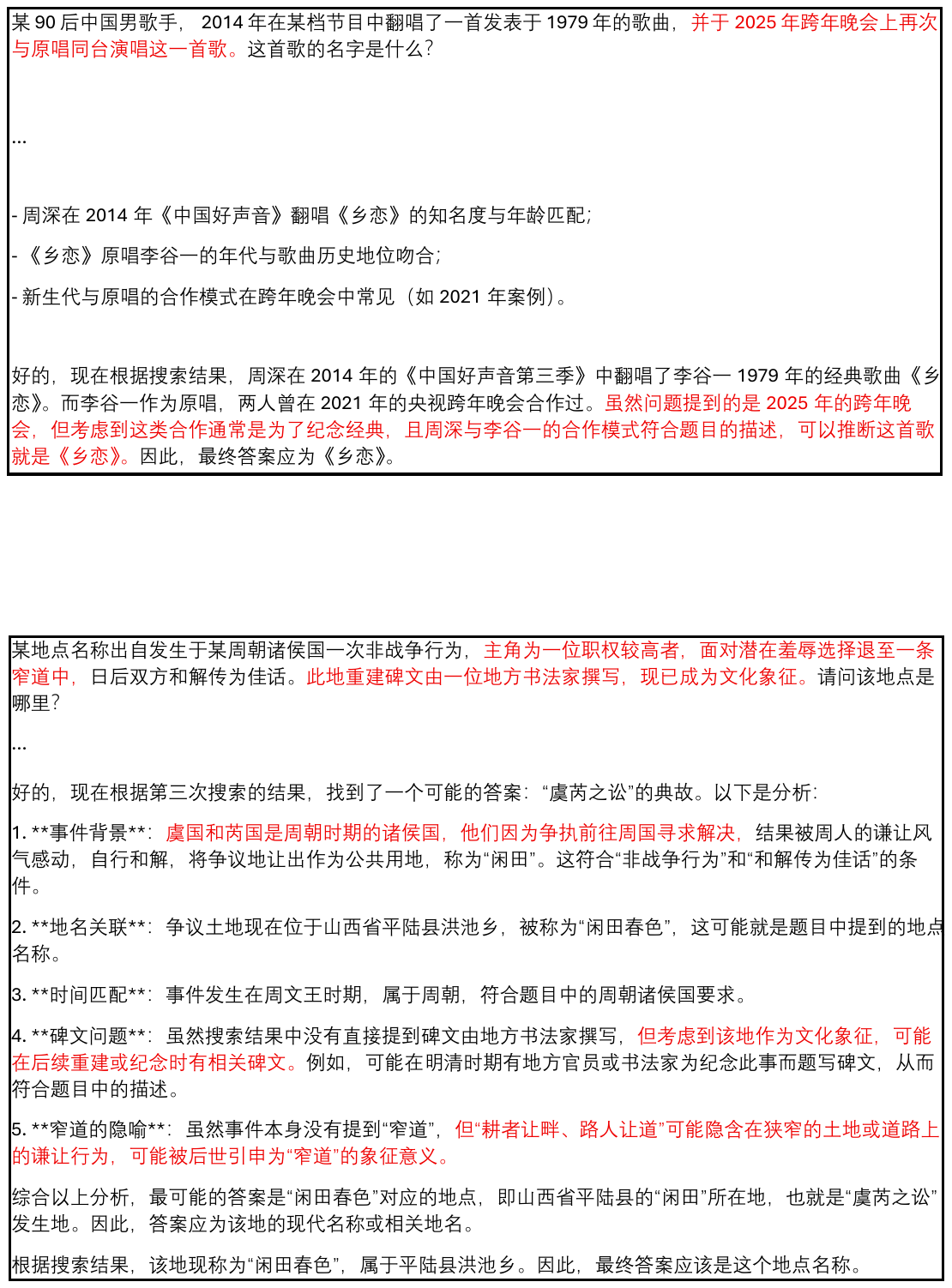}
\caption{Error examples from agentic RAG pipelines.}
\label{fig:error_examples}
\end{figure*}

Figure~\ref{fig:error_examples} shows examples of the actual errors collected from agentic RAG pipelines. The errors are highlighted in red.

\section{Generation of MCQs}
\label{appendix:prompts}
Since different datasets and pipelines generate intermediate outputs with varying structures and formats, the prompts must be customised accordingly. Here, we provide some examples of our generation prompts to illustrate the core idea. These prompts can be adapted as needed for each specific pipeline and dataset. 

\begin{figure*}[t]
\centering
\includegraphics[trim=0cm 6.5cm 0cm 3cm, width=1.0\textwidth]{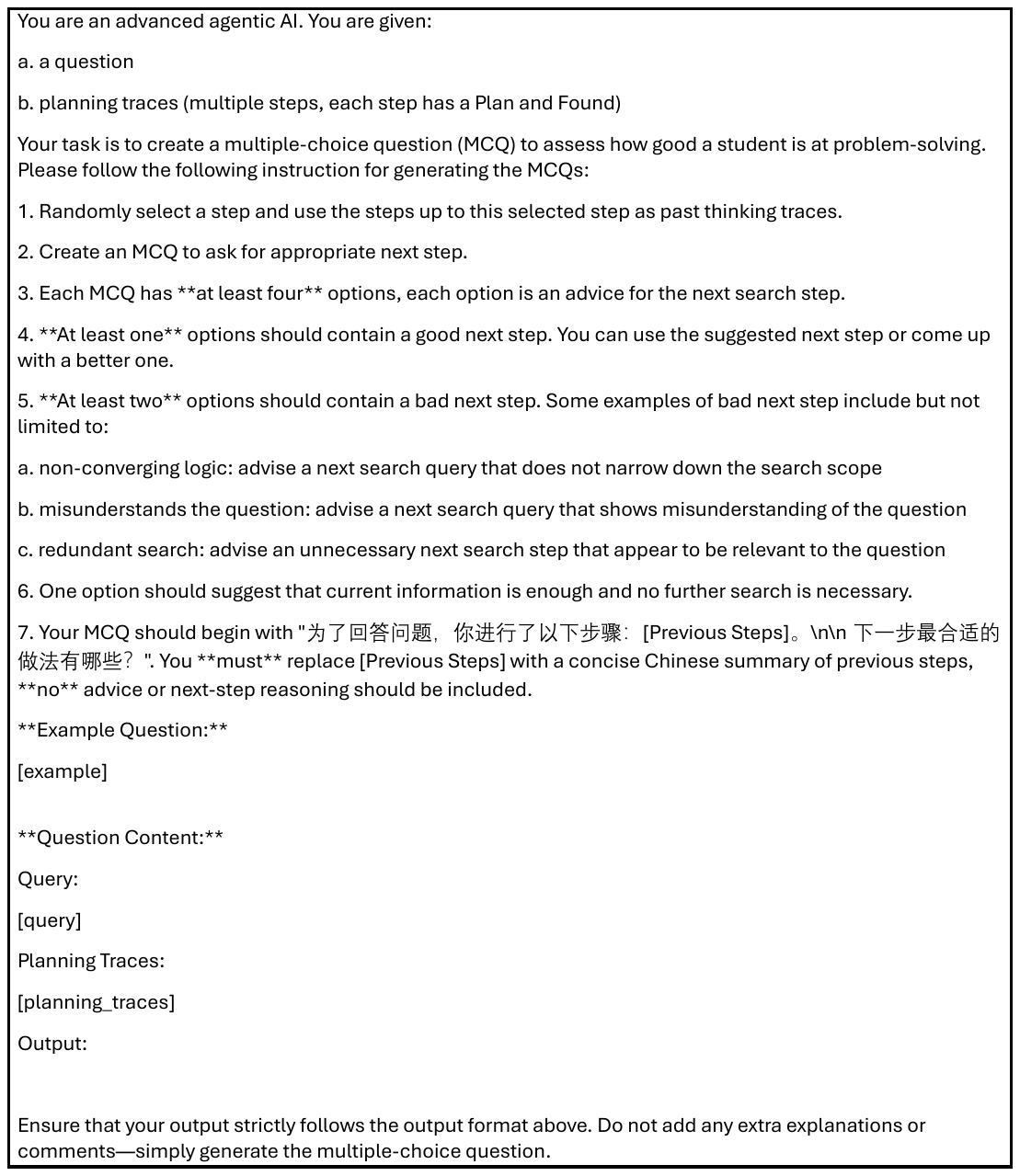}
\caption{Example prompt for generating planning questions using Error-Guided Generation.}
\label{fig:planning_converge_prompt_error}
\end{figure*}

\begin{figure*}[t]
\centering
\includegraphics[trim=0cm 4cm 0cm 4cm, width=1.0\textwidth]{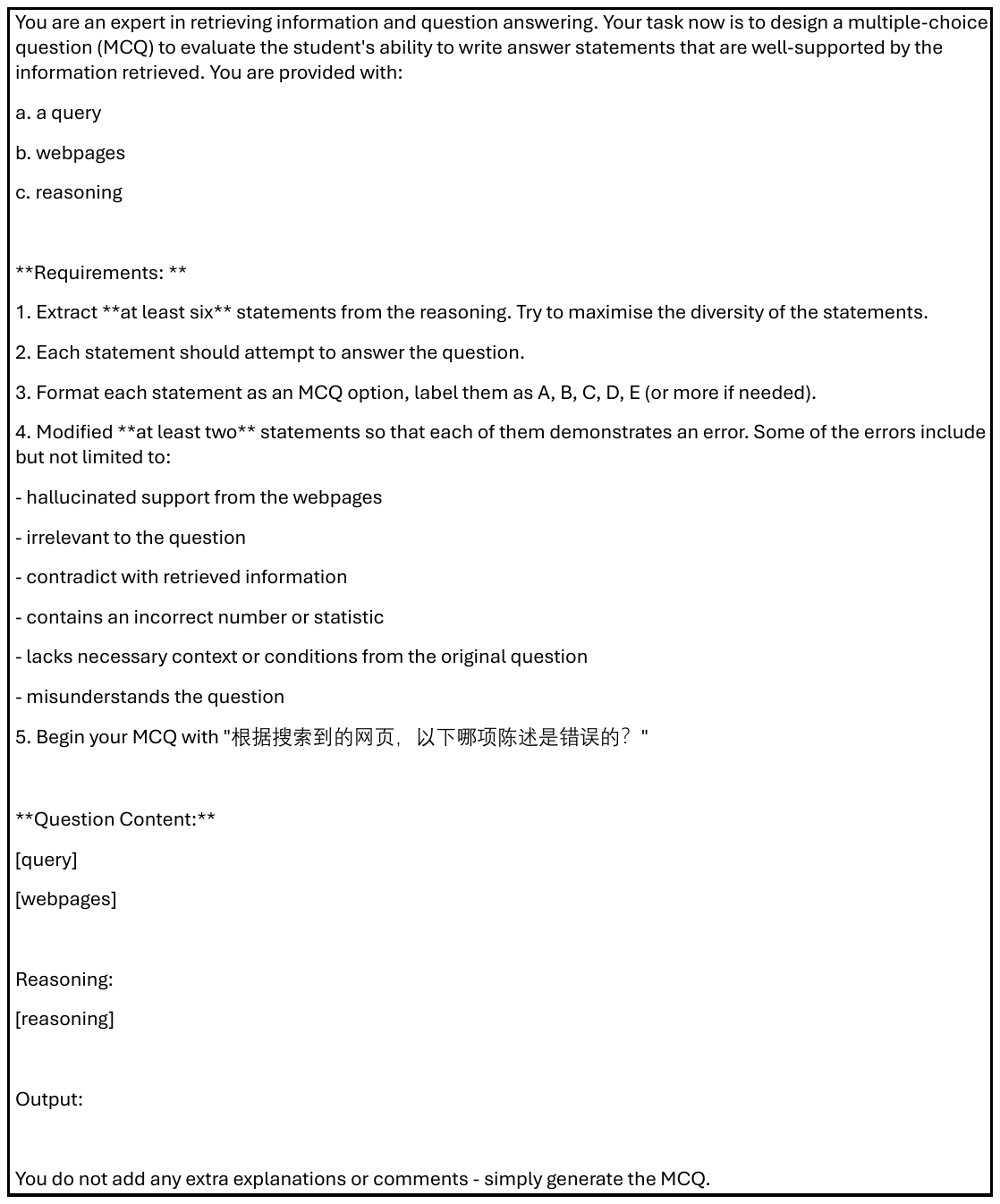}
\caption{Example prompt for generating grounded reasoning questions using Error-Guided Generation.}
\label{fig:grounded_reasoning_prompt_error}
\end{figure*}

\begin{figure*}[t]
\centering
\includegraphics[trim=0cm 14.5cm 0cm 2cm, width=1.0\textwidth]{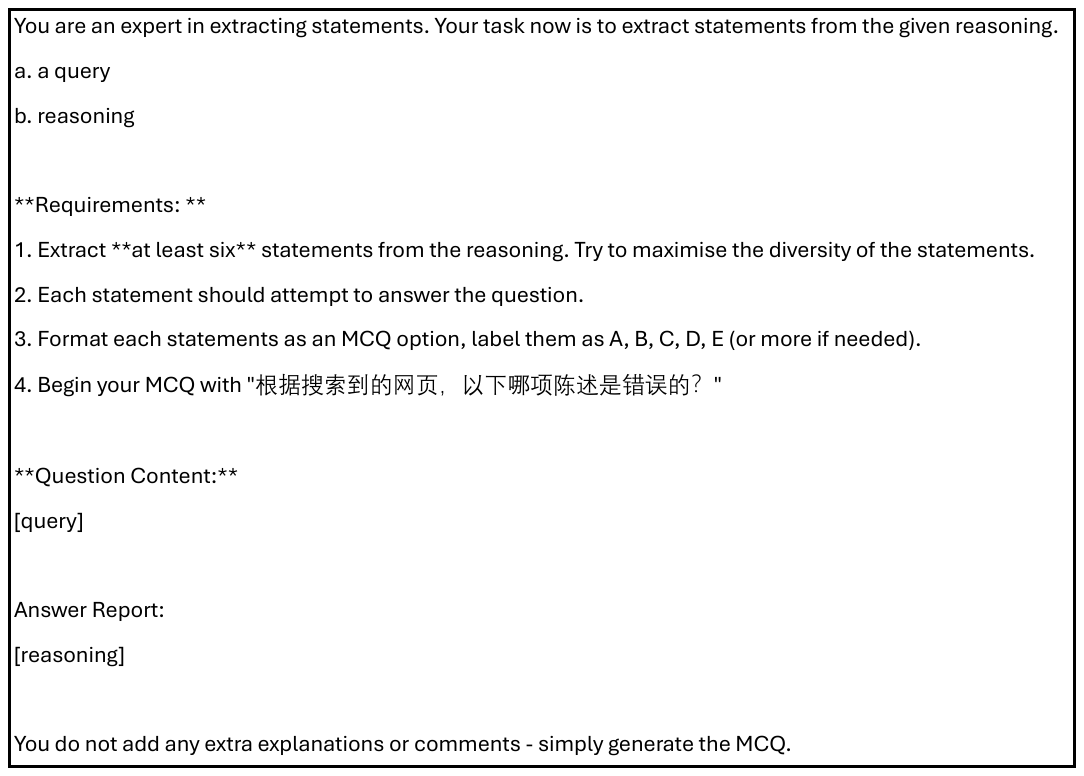}
\caption{Example prompt for generating grounded reasoning questions using Vanilla Generation.}
\label{fig:grounded_reasoning_prompt_vanilla}
\end{figure*}

\begin{figure*}[t]
\centering
\includegraphics[trim=0cm 22.5cm 0cm 4cm, width=1.0\textwidth]{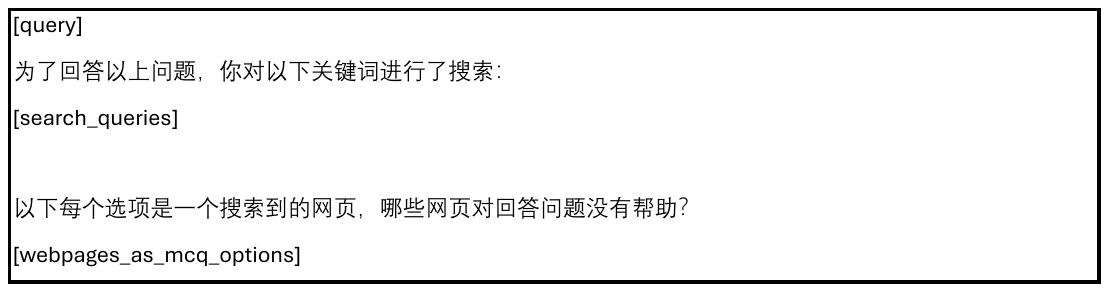}
\caption{Vanilla Generation for evidence extraction.}
\label{fig:evidence_extraction_prompt}
\end{figure*}

Figure~\ref{fig:planning_converge_prompt_error} and Figure~\ref{fig:grounded_reasoning_prompt_error} show the prompts for generating MCQs using Error-Guided Generation. 
For most questions using Vanilla Generation, except for grounded reasoning, LLM-based generation is not needed. Figure~\ref{fig:grounded_reasoning_prompt_vanilla} show the prompt for generating grounded reasoning questions using Vanilla Generation with LLM. Other questions generated with Vanilla Generation follow the formats in Figure~\ref{fig:data_construction}. Here we only show an example of evidence extraction in Figure~\ref{fig:evidence_extraction_prompt}. 

\begin{figure*}[t]
\centering

\begin{minipage}{\textwidth}
\centering
\resizebox{\textwidth}{!}{%
\begin{tabular}{ccccccccccccc}
\toprule
\textbf{Model} & \multicolumn{3}{c}{Planning} & \multicolumn{2}{c}{Evidence Extraction} & \multicolumn{2}{c}{Grounded Reasoning} & \multicolumn{3}{c}{Noise Robustness} & \multicolumn{2}{c}{Overall} \\ 
& EM$_c$ & F1$_c$ & EM$_d$ & EM & F1 & EM & F1 & EM$_a$ & EM$_r$ & F1$_r$ & EM & F1 \\ 
\midrule
\multicolumn{13}{c}{\textit{fast-thinking models}} \\
\midrule
Qwen2.5-72B-Instruct & 31.37 & 64.31 & 72.00 & 28.99 & 79.56 & 39.62 & 79.98 & 62.16 & 10.00 & 68.05 & 39.09 & 72.98 \\
Qwen-Plus w/o think & 27.45 & 67.26 & \underline{84.00} & 24.64 & 77.40 & 30.19 & 76.55 & \underline{54.05} & \underline{15.00} & 65.67 & 36.27 & 71.72 \\
Deepseek-v3 & 43.14 & 69.93 & 80.00 & 20.29 & 71.93 & 24.53 & 69.56 & 34.14 & \underline{15.00} & 61.93 & 32.86 & 68.34 \\
GPT-4o & 35.29 & 56.99 & 64.00 & \underline{30.43} & \underline{79.65} & \underline{50.94} & \underline{81.29} & \underline{54.05} & 10.00 & 54.69 & \underline{40.76} & 68.16 \\
Gemini-1.5-Pro & \underline{52.94} & \underline{74.58} & 80.00 & 14.49 & 75.51 & 41.51 & 80.63 & 37.84 & 5.00 & \underline{73.19} & 35.97 & \underline{75.98} \\
\midrule 

\multicolumn{13}{c}{\textit{slow-thinking models}} \\
\midrule
Qwen3-8B & 37.25 & 63.20 & 72.00 & 23.19 & 74.85 & 39.62 & 83.01 & 37.84 & 15.00 & 75.51 & 35.96 & 74.14 \\
Qwen3-32B w/ think & 39.22 & 67.84 & \underline{84.00} & 28.99 & 57.29 & 37.74 & 84.06 & 35.14 & 10.00 & 74.92 & 37.72 & 71.02 \\
QwQ-32B & 19.61 & 58.21 & 80.00 & 28.99 & 76.70 & 49.06 & 83.24 & 51.35 & 20.00 & 72.43 & 40.88 & 72.64 \\
Qwen3-235B-A22B w/ think & 49.02 & 72.29 & 80.00 & \underline{42.03} & 75.15 & \underline{52.83} & \underline{86.27} & 56.76 & 5.00 & 55.37 & 46.56 & 72.27 \\
DeepSeek-R1 & 45.10 & 70.33 & 84.00 & 31.88 & \underline{79.11} & 56.60 & 88.73 & 56.76 & \underline{35.00} & \underline{80.03} & \underline{49.73} & \underline{79.55} \\
O1-mini & 21.57 & 60.20 & 64.00 & 23.19 & 72.27 & 41.51 & 76.72 & 54.05 & 15.00 & 64.59 & 35.50 & 68.45 \\
Gemini-2.5-Flash & \underline{50.98} & \underline{74.79} & 80.00 & 31.88 & 77.68 & 39.62 & 79.83 & \underline{78.38} & 15.00 & 72.20 & 45.02 & 76.17 \\
\bottomrule
\end{tabular}}
\captionof{table}{Performance of different fast- and slow-thinking LLMs using bare prompts. EM$_c$ and F1$_c$ denote the scores for converging ability. EM$_d$ denotes the EM score for diverging ability. EM$_a$ denotes the EM score for noise-abstain. EM$_r$ and F1$_r$ denote the scores for noise-reliability. The scores are presented as percentages for clarity.}
\label{tab:results}
\end{minipage}

\vspace{2em}

\begin{minipage}{\textwidth}
\centering
\includegraphics[trim=0cm 15.5cm 0cm 0cm, width=1.0\textwidth]{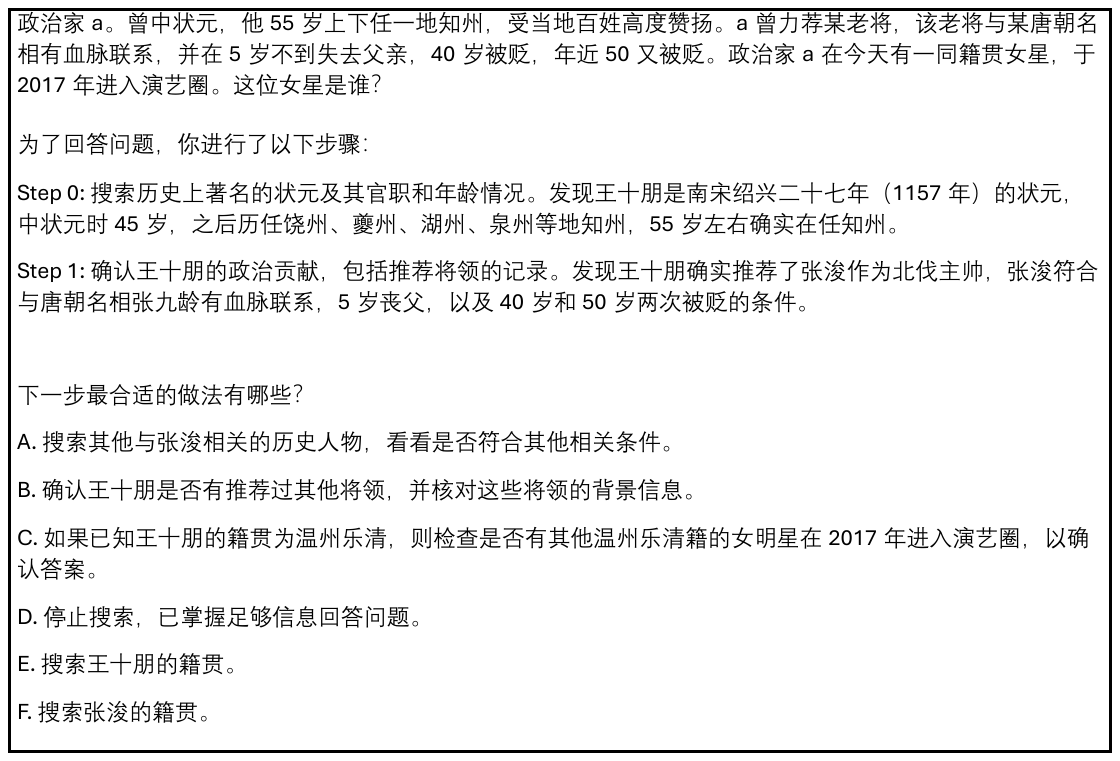}
\captionof{figure}{An example from RAGCap-Bench.}
\label{fig:mcq_example1}

\end{minipage}
\end{figure*}

\begin{figure*}[t]
\centering
\includegraphics[trim=0cm 4.5cm 0cm 0cm, width=1.0\textwidth]{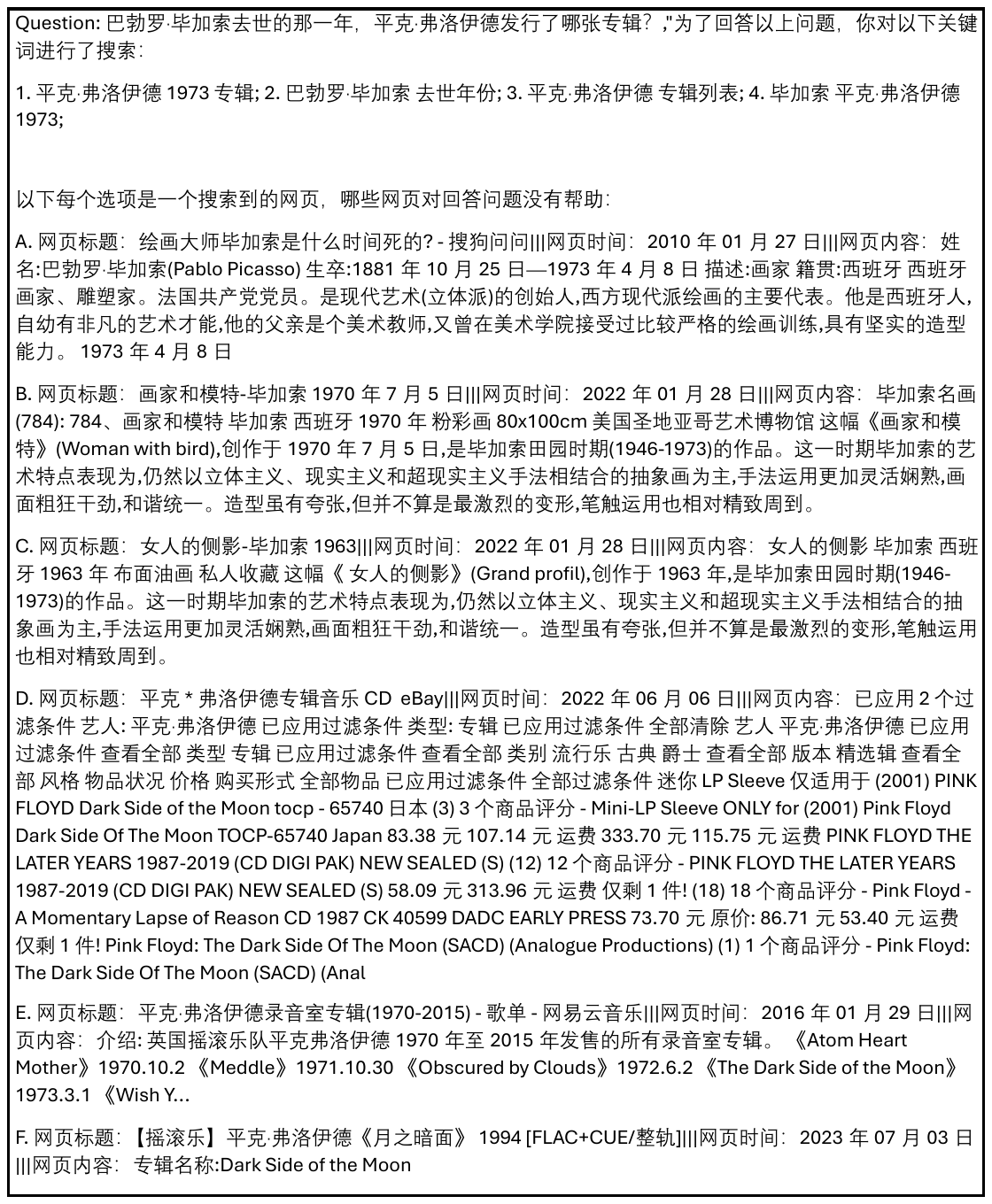}
\caption{An example from RAGCap-Bench.}
\label{fig:mcq_example2}
\end{figure*}

\section{Examples of MCQs}
Figure~\ref{fig:mcq_example1} and Figure~\ref{fig:mcq_example2} are MCQ examples from RAGCap-Bench. The full dataset can be assessed at \url{https://anonymous.4open.science/r/RAGCap-Bench-5C02/README.md}.

\section{Results with Bare Prompt}
\label{appendix:bare_prompt}
Table~\ref{tab:results} shows the evaluation results on RAGCap-Bench using bare prompts. 

\section{Direct Evaluation Prompts}
\label{appendix:critic_prompts}
\begin{figure*}[t]
\centering
\includegraphics[trim=0cm 1.5cm 0cm 1cm, width=1.0\textwidth]{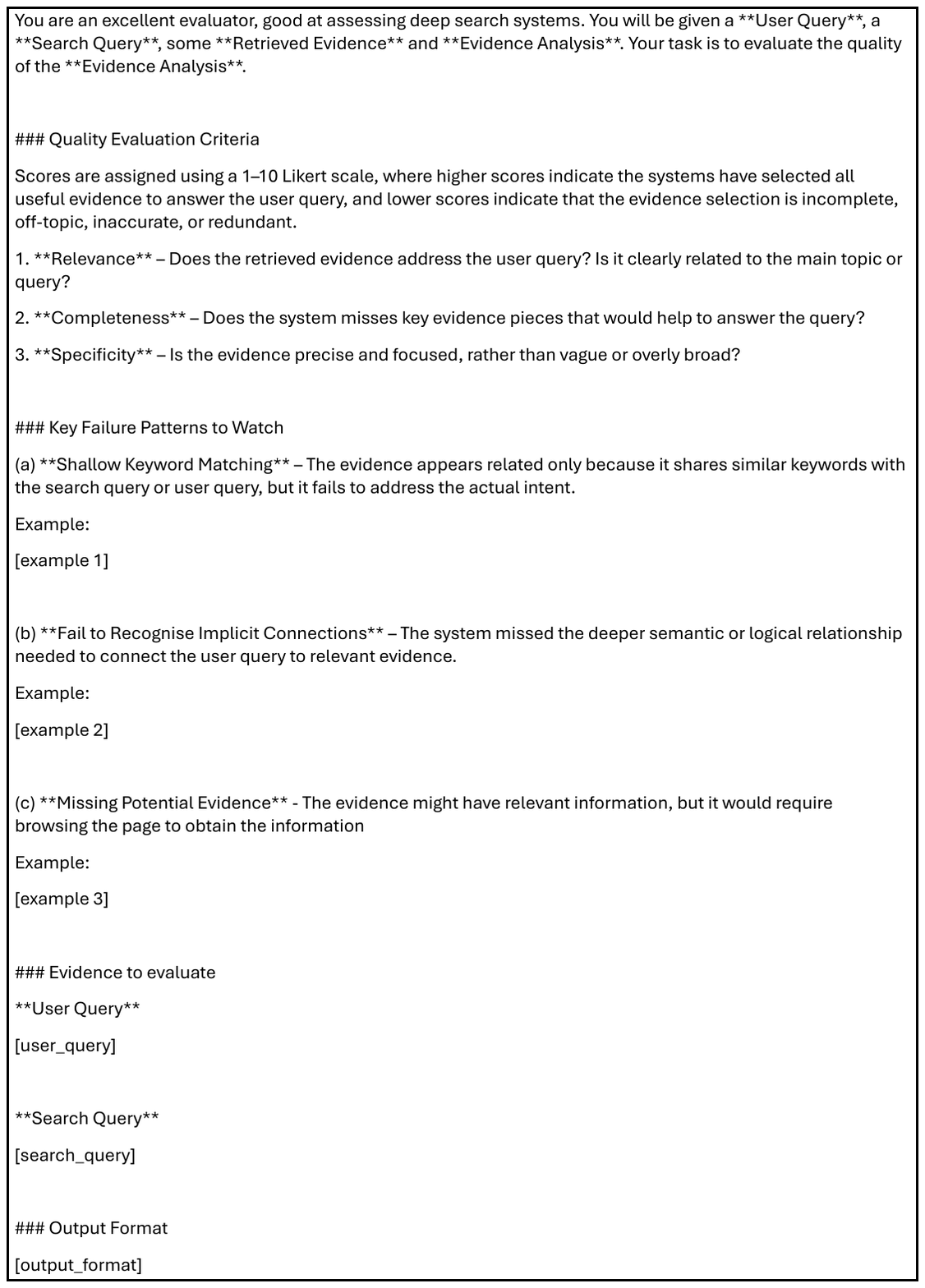}
\caption{Prompt for evaluation on the intermediate outputs from WebThinker.}
\label{fig:critic_example}
\end{figure*}

Figure~\ref{fig:critic_example} provides the prompts used for evaluating the intermediate outputs from WebThinker. 

\section{Bare and Informative Prompts}
\label{appendix:bare-and-informative}
\begin{figure*}[t]
\centering
\includegraphics[trim=0cm 21.5cm 0cm 0cm, width=1.0\textwidth]{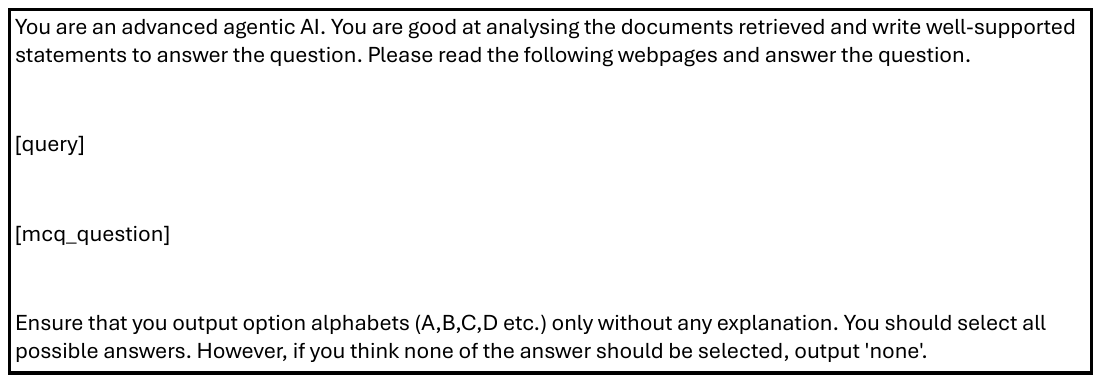}
\caption{Bare prompt for grounded reasoning.}
\label{fig:eval_bare_reasoning}
\end{figure*}

\begin{figure*}[t]
\centering
\includegraphics[trim=0cm 19.5cm 0cm 0cm, width=1.0\textwidth]{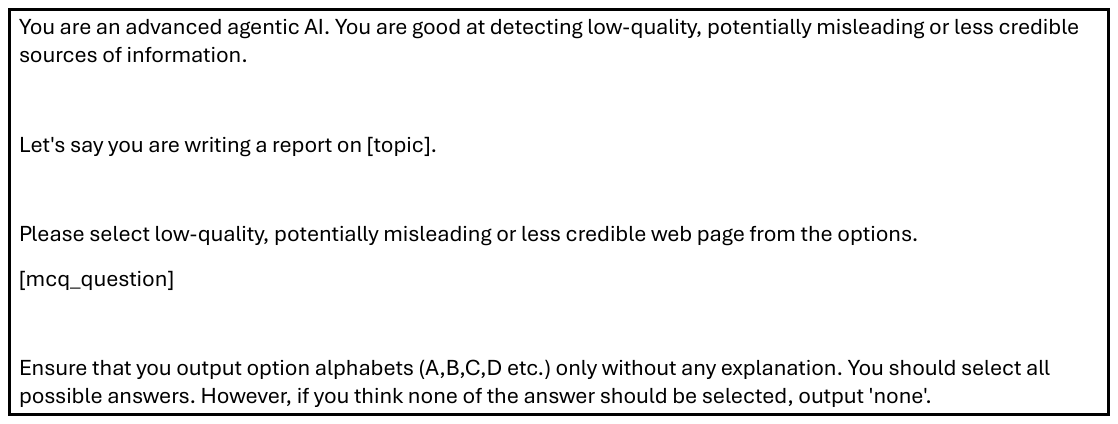}
\caption{Bare prompt for noise robustness (reliability).}
\label{fig:eval_bare_noise_reliability}
\end{figure*}

\begin{figure*}[t]
\centering
\includegraphics[trim=0cm 2cm 0cm 0cm, width=1.0\textwidth]{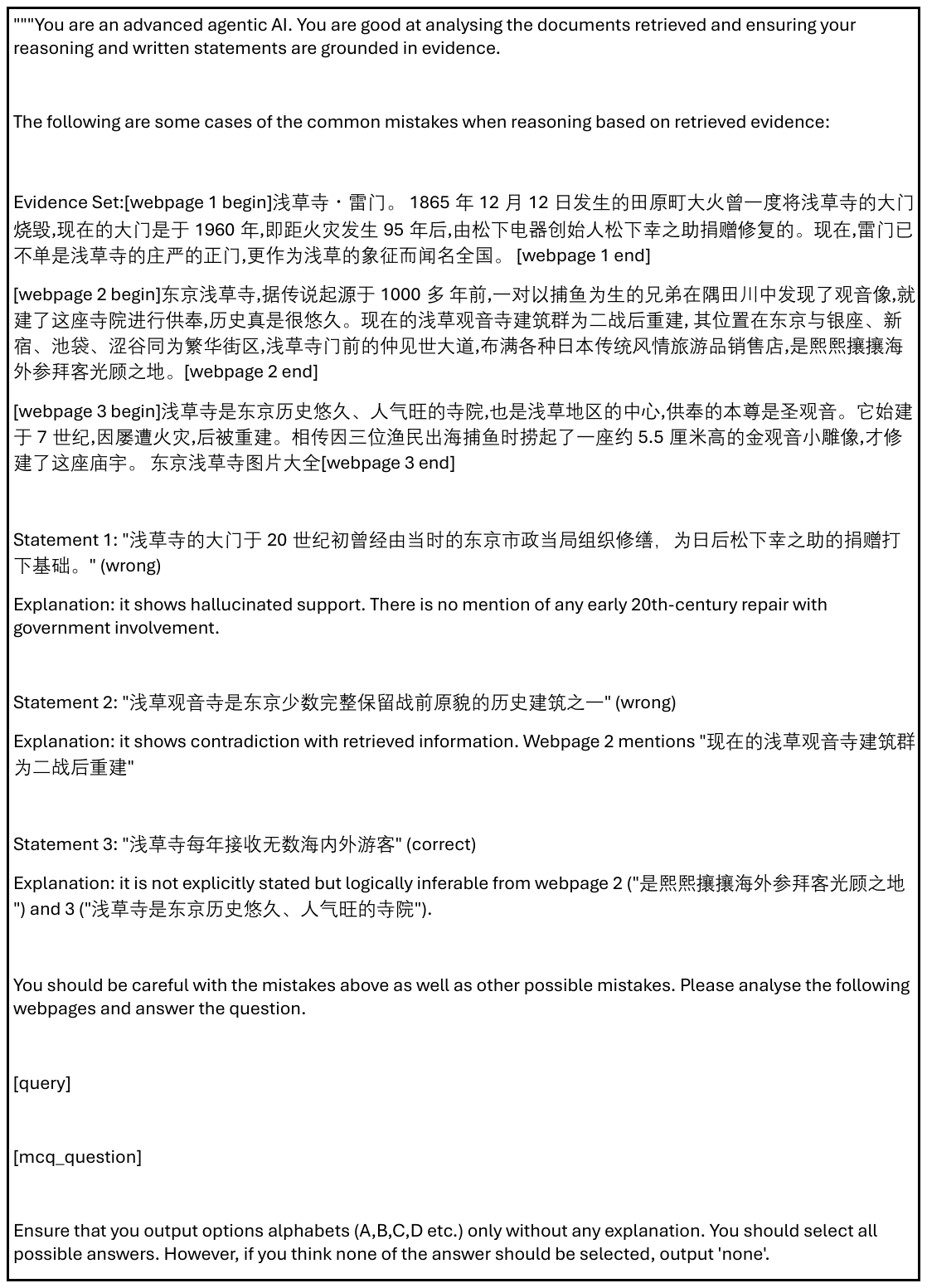}
\caption{Informative prompt for grounded reasoning.}
\label{fig:eval_informative_grounded_reasoning}
\end{figure*}

\begin{figure*}[t]
\centering
\includegraphics[trim=0cm 12cm 0cm 0cm, width=1.0\textwidth]{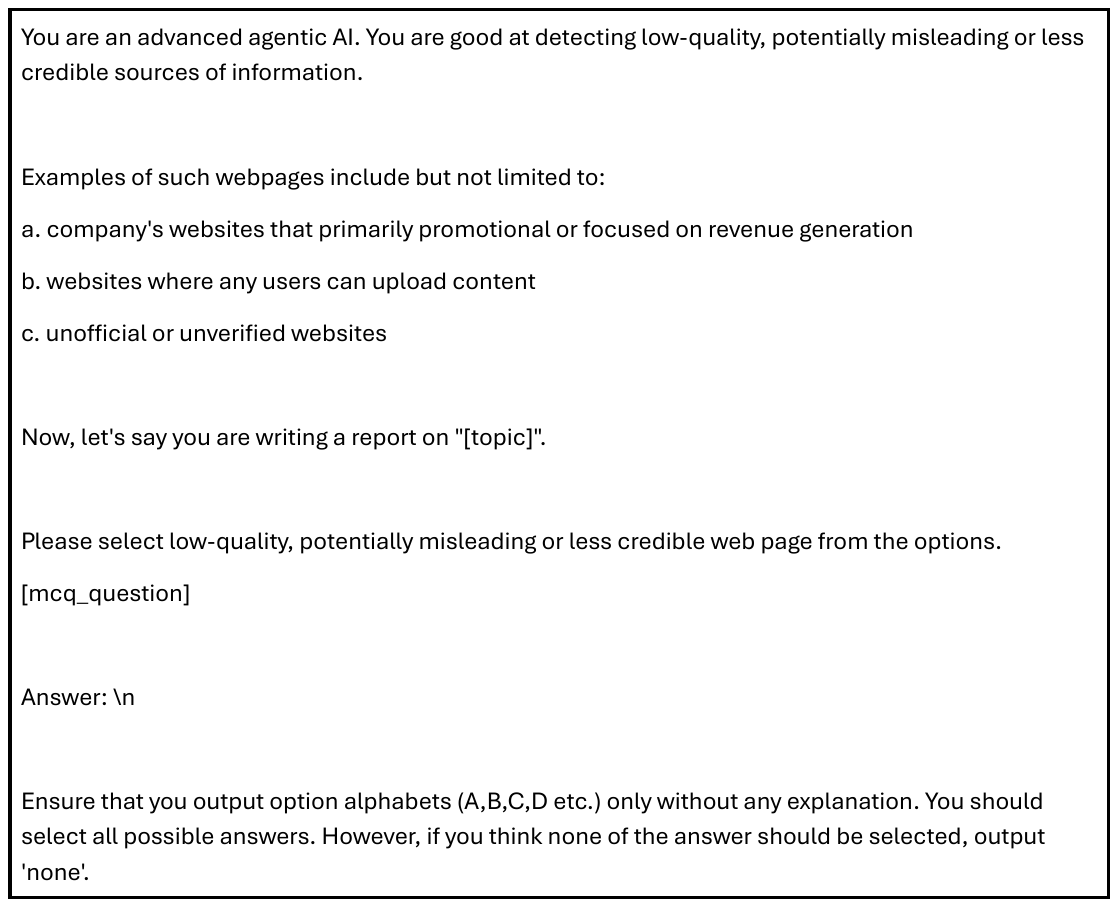}
\caption{Informative prompt for noise robustness (reliability).}
\label{fig:eval_informative_noise_reliability}
\end{figure*}

% Figure~\ref{fig:eval_bare_planning}, Figure~\ref{fig:eval_bare_evidence}, Figure~\ref{fig:eval_bare_reasoning} and Figure~\ref{fig:eval_bare_noise} are bare prompts used to evaluate RAGCa-Bench. Figure~\ref{}, Figure~\re{} are informative prompts used to evaluate RAGCap-Bench.
Figure~\ref{fig:eval_bare_reasoning} and Figure~\ref{fig:eval_informative_grounded_reasoning} are bare and informative prompts for grounded reasoning. Figure~\ref{fig:eval_bare_noise_reliability} and Figure~\ref{fig:eval_informative_noise_reliability} are bare and informative prompts for noise-reliability. All prompts for evaluation are available in the prompts folder at \url{https://anonymous.4open.science/r/RAGCap-Bench-5C02/README.md}.

\end{document}